\documentclass[journal]{IEEEtran}

\usepackage{fancyvrb}
\usepackage[pdftex]{graphicx}
\usepackage{hyperref}
\usepackage{ragged2e}
\usepackage[section]{placeins}
\usepackage{cite}
\usepackage{tipa}
\usepackage{wasysym}
\usepackage{gensymb}

\begin{document}

\title{Autonomous Robots and the SP Theory of Intelligence}

\author{J Gerard Wolff,~\IEEEmembership{Member,~IEEE}%
\thanks{Dr Gerry Wolff is founder and director of CognitionResearch.org, Menai Bridge, UK. e-mail: jgw@cognitionresearch.org.}%
\thanks{Manuscript received [Month] [day], [year]; revised [Month] [day], [year].}}

\markboth{IEEE Access,~Vol.~X, No.~Y, [month]~[year]}%
{Wolff: Autonomous Robots and the SP Theory of Intelligence}

\maketitle

\begin{abstract}

This paper is about how the {\em SP theory of intelligence} and its realisation in the {\em SP machine} (both outlined in the paper) may help in the design of the `brains' of autonomous robots, meaning robots that do not depend on external intelligence or power supplies, are mobile, and have human-like versatility and adaptability in intelligence. The paper addresses three main problems: how to increase the computational and energy efficiency of computers, and to reduce their size and weight; how to achieve human-like versatility in intelligence; and likewise for human-like adaptability in intelligence. Regarding the first problem, the SP system has potential for substantial gains in computational efficiency, with corresponding cuts in energy consumption and in the bulkiness of computers: by reducing the size of data to be processed; by exploiting statistical information that the system gathers as an integral part of how it works; and via a new version of Donald Hebb's concept of a ``cell assembly''. Towards human-like versatility in intelligence, the SP system has strengths in unsupervised learning, natural language processing, pattern recognition, information retrieval, several kinds of reasoning, planning, problem solving, and more, with seamless integration amongst structures and functions. The SP system's strengths in unsupervised learning and other aspects of intelligence may help to achieve human-like adaptability in intelligence via: 1) one-trial learning; 2) the learning of natural language; 3) learning to see; 4) building 3D models of objects and of a robot's surroundings; 5) learning regularities in the workings of a robot and in the robot's environment; 6) exploration and play; 7) learning major skills; and 7) learning via demonstration. Also discussed are: how the SP system may process parallel streams of information; generalisation of knowledge, correction of over-generalisations, and learning from dirty data; how to cut the cost of learning; and reinforcements and motivations.

\end{abstract}


\section{Introduction}

This paper is about how the {\em SP theory of intelligence} and its realisation in the {\em SP machine} (both of them to be described) may help in the design of the information-processing `brains' of autonomous robots.\footnote{In the rest of this paper, the quote marks will be omitted when referring to the information-processing mechanisms or brains of autonomous robots.}\textsuperscript{,}\footnote{As in \cite[Note 21]{sp_benefits_apps}, this paper does not in any way endorse or defend the unethical or illegal use of autonomous robots of any kind to cause death, injury, or damage to property.} Here, `autonomous robots' are ones that do not depend on external intelligence (natural or artificial), do not depend on external power supplies, and are mobile. We shall also assume that a goal in their development is to provide them with human-like versatility and adaptability in intelligence.

The paper is relevant to robots that are not autonomous in the sense just described, but the problems to be addressed are most acute in robots that are intended to function autonomously, and potential solutions are correspondingly more interesting.

In brief, the problems and potential solutions to be discussed are:

\begin{itemize}

\item {\em Computational efficiency, the use of energy, and the size and weight of computers}. If a robot is to be autonomous in the sense outlined above, it needs a brain that is efficient enough to do all the necessary processing without external assistance, does not require an industrial-scale power station to meet its energy demands, and is small enough and light enough to be carried around easily---things that are difficult or impossible to achieve with current technologies.

    The SP system may help: by reducing the size of data to be processed; by exploiting statistical information that the system gathers as an integral part of how it works; and via a new version of Donald Hebb's \cite{hebb_1949} concept of a ``cell assembly''. % These possibilities are discussed in Section \ref{efficiency_energy_bulk_section}.

\item {\em Towards human-like versatility in intelligence}. If a robot is to operate successfully in an environment where people cannot help, or where such opportunities are limited, it needs as much as possible of the versatility in intelligence that people may otherwise provide.

    The SP system demonstrates versatility via its strengths in areas such as unsupervised learning, natural language processing, pattern recognition, information retrieval, several kinds of reasoning, planning, problem solving, and more.

    But the SP system is not simply a kludge of different AI functions. Owing to its focus on simplification and integration of concepts in computing and cognition (Section \ref{outline_of_sp_system}), it promises to reduce or eliminate unnecessary complexity and to avoid awkward incompatibilities between poorly-integrated subsystems. And like any theory that simplifies and integrates a good range of observations and concepts, it promises deeper insights and better solutions to problems than may otherwise be achieved.

    % How the SP system may help to achieve human-like versatility in intelligence in autonomous robots is discussed in Section \ref{towards_versatility_section}.

\item {\em Towards human-like adaptability in intelligence}. Amongst the AI capabilities of the SP system mentioned above, unsupervised learning has particular significance because of its potential as a key to human-like adaptability in intelligence, both directly and as a basis for other kinds of learning. % How unsupervised learning by the SP system may help to achieve human-like adaptability in the intelligence of autonomous robots is discussed in Section \ref{towards_adaptability_section}.

\end{itemize}

These problems and their potential solutions are discussed in Sections \ref{efficiency_energy_bulk_section}, \ref{towards_versatility_section}, and \ref{towards_adaptability_section}, below. As a foundation for those three sections, there is an outline of the SP theory in Section \ref{outline_of_sp_system}, with pointers to where further information may be found.

\subsection{Novelty and Contribution}

Like any good theory, the SP theory has a wide range of potential applications, some of which are described in \cite{sp_benefits_apps}. For reasons indicated above, and expanded in Sections \ref{efficiency_energy_bulk_section}, \ref{towards_versatility_section}, and \ref{towards_adaptability_section}, the SP theory is particularly relevant to the development of the brains of autonomous robots.

But mere assertion is not enough. This paper aims to explain why the SP theory is relevant to the development of robot brains and how it may be applied in that area. It is a considerable expansion of the short discussion in \cite[Section 6.3]{sp_benefits_apps}, with much new thinking, and it is substantially different from other publications in the SP programme of research (Section \ref{outline_of_sp_system}). To help make the paper comprehensible and readable, it includes summaries of material that has been presented more fully elsewhere. Unnecessary repetition of information has been minimised and there are frequent pointers to where more detailed information may be found.

The approach is radically different from other work in robotics because of its distinctive features (Section \ref{distinctive features_section}), especially the concept of multiple alignment (Section \ref{ma_section}). As will be seen in the body of the paper, this new approach has considerable potential in the three main areas that are addressed.

Autonomous robots have set a new challenge for the SP theory: how to process parallel streams of information. This has led to important new refinements of the theory, described in Sections \ref{interactions_regularities_section}, \ref{exploration_play_section}, and \ref{learning_skills_section}, and in Appendix \ref{parallel_streams_of_information_section}. The paper also includes new thinking about the identification of low-level perceptual features (Appendix \ref{identification_of_symbols_section}) and about quantification in the SP system (Appendix \ref{quantification_section}).

\section{Outline of the SP Theory and the SP Machine}\label{outline_of_sp_system}

In this section, and elsewhere in the paper, the aim is to describe the SP theory with enough detail to ensure that the rest of the paper makes sense.

The SP theory is a unique attempt to simplify and integrate concepts across artificial intelligence, mainstream computing, mathematics, and human perception and cognition, with information compression as a unifying theme.\footnote{The name ``SP'' is short for {\em Simplicity} and {\em Power}, because compression of any given body of information, {\bf I}, may be seen as a process of reducing informational “redundancy” in {\bf I} and thus increasing its ``simplicity'', whilst retaining as much as possible of its non-redundant expressive ``power''.}

The theory is described most fully in \cite{wolff_2006} and at some length in \cite{sp_extended_overview}. In addition to the present paper, potential benefits and applications of the theory are described in \cite{sp_benefits_apps} (an overview), \cite{sp_big_data} (how the SP system may help to solve problems associated with big data), \cite{sp_vision} (application of the SP theory to the understanding of natural vision and the development of computer vision), \cite{wolff_medical_diagnosis} (application of the SP system to medical diagnosis), and \cite{wolff_sp_intelligent_database} (the SP system as an intelligent database).

\subsection{The SP Computer Model and the SP Machine}\label{computer_model_and_sp_machine_section}

The SP theory is realised in the form of a computer model, SP70, which may be regarded as a first version of the {\em SP machine}.

Expressing the theory in the form of a computer model helps to reduce vagueness in the theory. Perhaps more importantly, it provides a means of testing candidate ideas. As a result of such testing, many seemingly-promising ideas have been rejected. The model is also a means of demonstrating what can be achieved with the system.

An outline of how the SP computer model works may be found in \cite[Section 3.9]{wolff_2006}, with more detail, including pseudocode, in \cite[Sections 3.10 and 9.2]{wolff_2006}.\footnote{The description of how the SP70 model works includes a description, in \cite[Sections 3.9.1 and 3.10]{wolff_2006}, of a subset of the SP70 model called SP61.} Fully commented source code for the SP70 computer model may be downloaded via a link near the bottom of \href{http://www.cognitionresearch.org/sp.htm}{www.cognitionresearch.org/sp.htm}.

\begin{figure}[!htbp]
\centering
\includegraphics[width=0.45\textwidth]{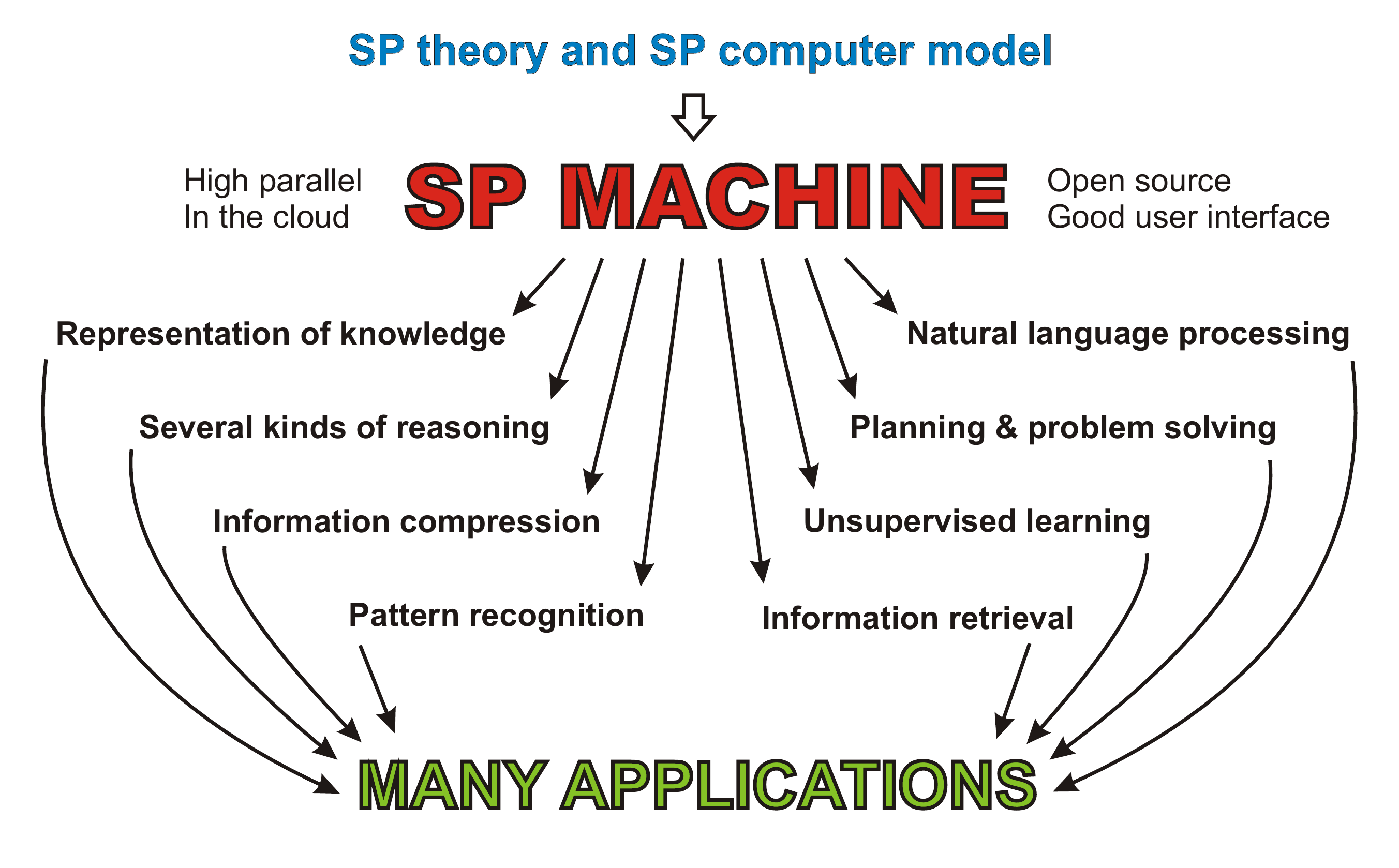}
\caption{Schematic representation of the development and application of the SP machine. Reproduced from Figure 2 in \cite{sp_extended_overview}, with permission.}
\label{sp_machine_figure}
\end{figure}

It is envisaged that the SP computer model will be the basis for the creation of a high-parallel, open-source version of the SP machine, hosted on an existing high-performance computer. This will be a means for researchers everywhere to explore what can be done with the system and to create new versions of it \cite[Section 3]{sp_benefits_apps}, \cite[Section XII]{sp_big_data}. How things may develop is shown schematically in Figure \ref{sp_machine_figure}.

\subsection{Patterns and Symbols}\label{patterns_and_symbols_section}

In the SP system, knowledge is represented with arrays of atomic {\em symbols} in one or two dimensions called {\em patterns}. The SP70 computer model works with 1D patterns but it is envisaged that the system will be generalised to work with 2D patterns \cite[Section 3.3]{sp_extended_overview}.

A `symbol' in the SP system is simply a mark that can be matched with any other symbol to determine whether it is the same or different: no other result is permitted.

With one exception, any meaning associated with a given SP symbol or combination of symbols must be expressed using other SP symbols. The exception is where an SP symbol connects with an entity or value outside the SP system. For example, a signal from a sensor in an autonomous robot, or an instruction for one of the robot's muscles to contract, may be represented by a symbol within the SP system.\footnote{It is pertinent here to mention that some symbols are classified as `identification' or `ID' symbols, while others are classified as `contents' or `C' symbols \cite[Section 3.4.5]{wolff_2006}. But these distinctions serve the internal workings of the SP system and do not impinge directly on how the system functions in the representation and processing of knowledge.}

Although conventional computing systems make extensive use of numbers, the SP system, in itself, makes no provision for the representation or processing of numbers. Possible responses to this feature of the SP system are discussed in Appendix \ref{quantification_section}.

The way in which SP symbols may be identified in such things as images, speech, and music, is discussed in Appendix \ref{identification_of_symbols_section}.

In themselves, SP patterns are not particularly expressive. But within the multiple alignment framework (Section \ref{ma_section}), they support the representation and processing of a wide variety of kinds of knowledge (Section \ref{ma_knowledge_section}). A goal of the SP research programme is to establish one system for the representation and processing of {\em all} kinds of knowledge (see also \cite[Section III]{sp_big_data}). Evidence to date suggests that this may be achieved with SP patterns in the multiple alignment framework.

Any collection of SP patterns is termed a {\em grammar}. Although that term is most closely associated with linguistics, it will be used throughout this paper for a collection of SP patterns describing any kind of knowledge.\footnote{This is partly because research on grammatical inference is one of the inspirations for the SP concepts (Section \ref{computer_models_of_language_learning_section}), and partly because of the significance of grammars in research on principles of minimum length encoding.}

\subsection{Information Compression}\label{information_compression_section}

The SP theory is conceived as a brain-like system that receives {\em New} information via its senses and stores some or all of it in compressed form as {\em Old} information. The emphasis on information compression derives from earlier research on grammatical inference (Section \ref{computer_models_of_language_learning_section}) and the principle of {\em minimum length encoding} (MLE) \cite{solomonoff_1964,wallace_boulton_1968,rissanen_1978}).

At an abstract level, information compression means the detection and reduction of {\em redundancy} in information. In more concrete terms, redundancy means {\em recurrent patterns}, {\em regularities}, {\em structures}, and {\em associations}, including {\em causal associations}. Thus information compression provides a means of discovering such things as words in natural language (Section \ref{computer_models_of_language_learning_section}), objects (Section \ref{building_model_of_world_section}), and the association between lightning and thunder \cite[Section III-A.1]{sp_big_data}, in accordance with the DONSVIC principle \cite[Section 5.2]{sp_extended_overview}.\footnote{{\em DONSVIC} = ``The discovery of natural structures via information compression''.}

The default assumption in the SP theory is that compression of information is always lossless, meaning that all non-redundant information is retained. In particular applications, there may be a case for discarding non-redundant information (see, for example, \cite[Section X-B]{sp_big_data}) but any such discard is reversible.

In the SP system, information compression is achieved via the matching and unification of patterns. More specifically, it is achieved via the building of multiple alignments and via the unsupervised learning of grammars. These three things are described briefly in the following three subsections.

\subsubsection{Information Compression Via the Matching and Unification of Patterns}\label{mup_section}

The basis for information compression in the SP system is a process of searching for patterns that match each other with a process of merging or `unifying' patterns that are the same. At the heart of the SP70 computer model is a method for finding good full and partial matches between sequences with advantages compared with classical methods \cite[Appendix A]{wolff_2006}.\footnote{The main advantages are \cite[Section 3.10.3.1]{wolff_2006}: 1) That it can match arbitrarily long sequences without excessive demands on memory; 2) For any two sequences, it can find a set of alternative matches (each with a measure of how good it is) instead of a single `best' match; 3) The `depth' or thoroughness of the searching can be controlled by parameters.}

\subsubsection{Information Compression Via the Building of Multiple Alignments}\label{ma_section}

That process for finding good full and partial matches between sequences is the foundation for processes that build {\em multiple alignments} like the one shown in Figure \ref{class_part_plant_figure}.\footnote{The concept of multiple alignment in the SP system has been borrowed and adapted from that concept in bioinformatics.}

\begin{figure*}[!htbp]
\fontsize{07.00pt}{08.40pt}
\centering
{\bf
\begin{BVerbatim}
0                 1                2                  3              4              5                  6

                  <species>
                  acris
                  <genus> ---------------------------------------------------------------------------- <genus>
                  Ranunculus ------------------------------------------------------------------------- Ranunculus
                                                                                    <family> --------- <family>
                                                                                    Ranunculaceae ---- Ranunculaceae
                                                                     <order> ------ <order>
                                                                     Ranunculales - Ranunculales
                                                      <class> ------ <class>
                                                      Angiospermae - Angiospermae
                                   <phylum> --------- <phylum>
                                   Plants ----------- Plants
                                   <feeding>
has_chlorophyll ------------------ has_chlorophyll
                                   photosynthesises
                                   <feeding>
                                   <structure> ------ <structure>
                                                      <shoot>
<stem> ---------- <stem> ---------------------------- <stem>
hairy ----------- hairy
</stem> --------- </stem> --------------------------- </stem>
                  <leaves> -------------------------- <leaves>
                  compound
                  palmately_cut
                  </leaves> ------------------------- </leaves>
                                                      <flowers> ------------------- <flowers>
                                                                                    <arrangement>
                                                                                    regular
                                                                                    all_parts_free
                                                                                    </arrangement>
                  <sepals> -------------------------------------------------------- <sepals>
                  not_reflexed
                  </sepals> ------------------------------------------------------- </sepals>
<petals> -------- <petals> -------------------------------------------------------- <petals> --------- <petals>
                                                                                    <number> --------- <number>
                                                                                                       five
                                                                                    </number> -------- </number>
                  <colour> -------------------------------------------------------- <colour>
yellow ---------- yellow
                  </colour> ------------------------------------------------------- </colour>
</petals> ------- </petals> ------------------------------------------------------- </petals> -------- </petals>
                                                                                    <hermaphrodite>
<stamens> ------------------------------------------------------------------------- <stamens>
numerous -------------------------------------------------------------------------- numerous
</stamens> ------------------------------------------------------------------------ </stamens>
                                                                                    <pistil>
                                                                                    ovary
                                                                                    style
                                                                                    stigma
                                                                                    </pistil>
                                                                                    </hermaphrodite>
                                                      </flowers> ------------------ </flowers>
                                                      </shoot>
                                                      <root>
                                                      </root>
                                   </structure> ----- </structure>
<habitat> ------- <habitat> ------ <habitat>
meadows --------- meadows
</habitat> ------ </habitat> ----- </habitat>
                  <common_name> -- <common_name>
                  Meadow
                  Buttercup
                  </common_name> - </common_name>
                                   <food_value> ----------------------------------- <food_value>
                                                                                    poisonous
                                   </food_value> ---------------------------------- </food_value>
                                   </phylum> -------- </phylum>
                                                      </class> ----- </class>
                                                                     </order> ----- </order>
                                                                                    </family> -------- </family>
                  </genus> --------------------------------------------------------------------------- </genus>
                  </species>

0                 1                2                  3              4              5                  6
\end{BVerbatim}
}
\caption{The best multiple alignment created by the SP model, with a set of New patterns (in column 0) that describe some features of an unknown plant, and a set of Old patterns, including those shown in columns 1 to 6, that describe different categories of plant, with their parts and sub-parts, and other attributes. Reproduced from Figure 16 in \protect\cite{sp_extended_overview}, with permission.}
\label{class_part_plant_figure}
\end{figure*}

This example shows the best multiple alignment created by the SP computer model when a set of New patterns (in column 0)\footnote{Specifically, the New patterns in this example are `\texttt{has\_chlorophyll}' (a pattern with one symbol), `\texttt{<stem> hairy </stem>}', `\texttt{<petals> yellow </petals>}', `\texttt{<stamens> numerous </stamens>}', and `\texttt{<habitat> meadows </habitat>}'. The patterns in a set like that may be presented to the system and processed in any order.} is processed in conjunction with a set of pre-existing Old patterns like those shown in columns 1 to 6. Here, the multiple alignment is `best' because it is the one that achieves the most economical description of the New patterns in terms of the Old patterns. The way in which that description or `encoding' is derived from a multiple alignment is explained in \cite[Section 3.5]{wolff_2006} and \cite[Section 4.1]{sp_extended_overview}. Like all other kinds of knowledge, encodings derived from multiple alignments are recorded using SP patterns (Section \ref{patterns_and_symbols_section}).

This multiple alignment may be interpreted as the result of a process of recognition (Section \ref{pattern_recognition_section}). The New patterns represent the features of some unknown plant and the Old patterns in columns 1 to 6 represent candidate categories, at several levels of abstraction: species `Meadow Buttercup' (column 1), genus {\em Ranunculus} (column 6), family {\em Ranunculaceae} (column 5), and so on.

\subsubsection{Information Compression Via the Unsupervised Learning of Grammars}\label{us_grammars_section}

As outlined in \cite[Section 3.9.2]{wolff_2006} and \cite[Section 5.1]{sp_extended_overview}, and described more fully in \cite[Chapter 9]{wolff_2006}, the SP system may, without assistance from a ``teacher'' or anything equivalent, derive one or more plausible grammars from a body of New patterns, with minimum length encoding as a guiding principle. In that process, multiple alignment has a central role as a source of SP patterns for possible inclusion in any grammar (Section \ref{creating_candidate_patterns_section}).

\subsubsection{Heuristic Search}\label{heuristic_search_section}

Like most problems in artificial intelligence, each of the afore-mentioned problems---finding good full and partial matches between patterns, finding or constructing good multiple alignments, and inferring one or more good grammars from a body of data---is normally too complex to be solved by exhaustive search.

With intractable problems like these, it is often assumed that the goal is to find theoretically ideal solutions. But with these and most other AI problems, ``The best is the enemy of the good''. By scaling back one's ambitions and searching for ``reasonably good'' solutions, it is often possible to find solutions that are useful, and without undue computational demands.

As with other AI applications, and as with the building of multiple alignments in bioinformatics, the SP70 model uses heuristic techniques in all three cases mentioned above. This means searching for solutions in stages, with a pruning of the search tree at every stage, guided by measures of success \cite[Appendix A; Sections 3.9 and 3.10; Chapter 9]{wolff_2006}. With these kinds of techniques, acceptably good approximate solutions can normally be found without excessive computational demands and with ``big O'' values that are within acceptable limits.

\subsection{Multiple Alignment and the Representation and Processing of Diverse Kinds of Knowledge}\label{ma_knowledge_section}

\sloppy The expressive power of SP patterns within the multiple alignment framework derives in large part from the way that symbols in one pattern may serve as links to one or more other patterns or parts thereof. One of several examples in Figure \ref{class_part_plant_figure} is how the pair of symbols `\texttt{<family>~...~</family>}' in column 6 serves to identify the pattern `\texttt{<family> ...~Ranunculales ...~<hermaphrodite> ...~poisonous ...~</family>}' in column 5.

In the figure, these kinds of linkages between patterns mean that the unknown plant (with characteristics shown in column 0) may be recognised at several different levels within a hierarchy of classes: genus, family, order, class, and so on. Although it is not shown in the example, the system supports cross classification.

In the figure, the parts and sub-parts of the plant are shown in such structures as `\texttt{<shoot>}' (column 3), `\texttt{<flowers>}' (column 5), `\texttt{<petals>}' (column 6), and so on.

As in conventional systems for object-oriented design, the system provides for inheritance of attributes (Section \ref{reasoning_section}). But unlike such systems, there is smooth integration of class hierarchies and part-whole hierarchies, without awkward inconsistencies \cite[Section 4.2.1]{wolff_sp_intelligent_database}.

More generally, SP patterns within the multiple alignment framework provide for the representation and processing of a wide variety of kinds of knowledge including: the syntax and semantics of natural language; class hierarchies and part-whole hierarchies (as just described); networks and trees; entity-relationship structures; relational knowledge; rules and several kinds of reasoning; patterns and pattern recognition; images; structures in three dimensions; and procedural knowledge. There is a summary in \cite[Section III-B]{sp_big_data}, and more detail in Section \ref{towards_versatility_section}.

\subsection{Information Compression, Prediction, and Probabilities}\label{ic_prediction_probabilities_section}

Owing to the close connection between information compression and concepts of prediction and probability \cite{li_vitanyi_2009}, the SP system is fundamentally probabilistic. Each SP pattern has an associated frequency of occurrence and probabilities may be calculated for each multiple alignment and for any inference that may be drawn from any given multiple alignment.

\subsection{SP-Neural}\label{sp-neural_section}

Part of the SP theory is the idea, described most fully in \cite[Chapter 11]{wolff_2006}, that the abstract concept of a {\em pattern} in the SP theory may be realised more concretely in the brain with a collection of neurons in the cerebral cortex called a {\em pattern assembly}.

The word ``assembly'' has been adopted in this term because the concept is quite similar to Hebb's \cite{hebb_1949} concept of a {\em cell assembly}. The main difference is that the concept of pattern assembly is unambiguously explicit in proposing that the sharing of structure between two or more pattern assemblies is achieved by means of `references' from one structure to another, as described and discussed in \cite[Section 11.4.1]{wolff_2006}). Another difference relates to one-trial learning, as outlined in Section \ref{one_trial_learning_section}.

\begin{figure}[!htbp]
\centering
\includegraphics[width=0.45\textwidth]{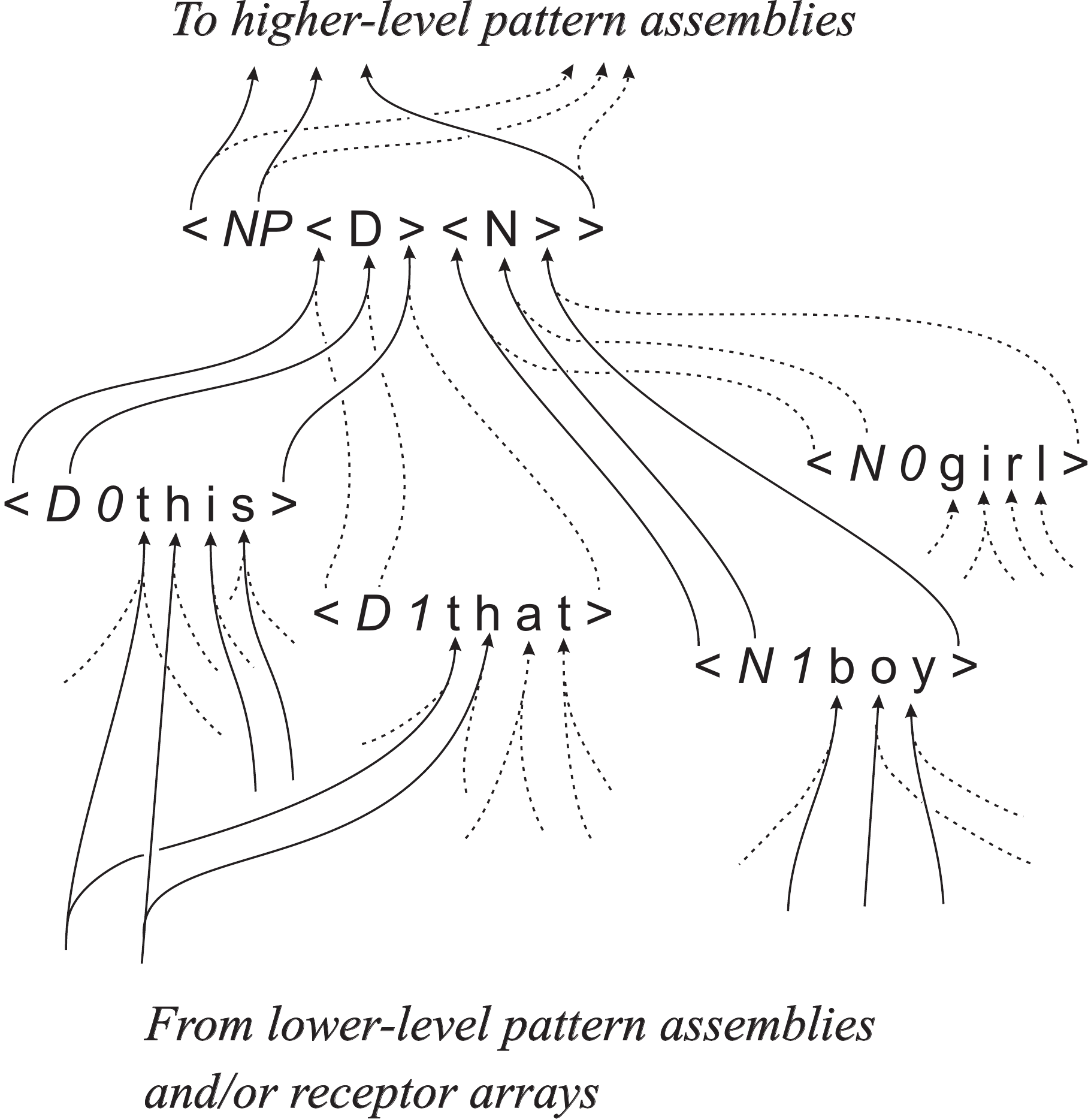}
\caption{Schematic representation of inter-connections amongst pattern assemblies as described in the text. Not shown in the figure are lateral connections within each pattern assembly, and inhibitory connections. Reproduced from Figure 11.2 in \cite{wolff_2006}, with permission.}
\label{connections_figure}
\end{figure}

Figure \ref{connections_figure} shows schematically how pattern assemblies may be represented and inter-connected with neurons. Here, each pattern assembly, such as `\texttt{< NP < D > < N > >}', is represented by the sequence of symbols of the corresponding SP pattern. Each symbol, such as `\texttt{<}' or `\texttt{NP}', would be represented in the pattern assembly by one neuron or a small group of inter-connected neurons.\footnote{As indicated in the caption to the figure, a more detailed representation would show lateral connections within each pattern assembly and inhibitory connections elsewhere. There is relevant discussion in \cite[Sections 11.3.3 and 11.3.4]{wolff_2006}.} Apart from the inter-connections amongst pattern assemblies, the cortex in SP-neural is somewhat like a sheet of paper on which knowledge may be written in the form of neurons.

It is envisaged that any pattern assembly may be `recognised' if it receives more excitatory inputs than rival pattern assemblies, perhaps via a winner-takes-all mechanism \cite[Section 11.3.4]{wolff_2006}. And, once recognised, any pattern assembly may itself be a source of excitatory signals leading to the recognition of higher-level pattern assemblies.

\subsection{Distinctive Features and Apparent Advantages of the SP Theory}\label{distinctive features_section}

Information compression and concepts of probability are themes in other research, including research on Bayesian inference, Kolmogorov complexity, deep learning, artificial neural networks, minimum length encoding, unified theories of cognition, natural language processing and more. The main features that distinguish the SP theory from these other areas of research, and apparent advantages compared with these other approaches, are:

\begin{itemize}

\item {\em Simplification and integration}. As mentioned above, the SP theory is a unique attempt to simplify and integrate concepts across artificial intelligence, mainstream computing, mathematics, and human perception and cognition:

\begin{itemize}

\item The canvass is much broader than it is, for example, in ``unified theories of cognition''. It has quite a lot to say, for example, about the nature of mathematics \cite{wolff_1993}, \cite[Chapter 10]{wolff_2006}.

\item In terms of achievement, not merely aspiration, the SP computer model combines simplicity with the ability to model a wide range of concepts and phenomena in computing and cognition (Section \ref{towards_versatility_section}).

\item The provision of one simple format for knowledge and one framework for the processing of knowledge promotes seamless integration of diverse structures and functions.

\end{itemize}

\item {\em The SP theory is a theory of computing}. Most other research is founded on the idea that computing may be understood in terms of the Universal Turing Machine or equivalent models such as Lamda Calculus or Post's Canonical System. By contrast, {\em the SP theory is itself a theory of computing} \cite[Chapter 4]{wolff_2006}.

\item {\em Intelligence}. What is distinctive about the SP theory as a theory of computing is that it provides much of the human-like intelligence that is missing from earlier models.\footnote{Although Alan Turing saw that computers might become intelligent \cite{turing_1950}, the Universal Turing Machine, in itself, does not tell us how! The SP theory, as it is now, goes some way towards plugging the gap, and has potential to do more.}

\item {\em Information compression via the matching and unification of patterns}. In trying to cut through the complexity of some other approaches, the SP research programme focusses on a simple, `primitive' idea: that information compression may be understood as a search for patterns that match each other, with the merging or `unification' of patterns that are the same (Section \ref{mup_section}).

\item {\em Multiple alignment}. More specifically, information compression via the matching and unification of patterns provides the basis for a concept of {\em multiple alignment}, outlined above (Section \ref{ma_section}). Developing this idea as a framework for the simplification and integration of concepts across a broad canvass has been a major undertaking. {\em Multiple alignment is a powerful and distinctive idea in the SP programme of research}.

\item {\em Transparency in the representation and processing of knowledge}. By contrast with sub-symbolic approaches to artificial intelligence, and notwithstanding objections to symbolic AI,\footnote{See, for example, ``Hubert Dreyfus's views on artificial intelligence'', {\em Wikipedia}, \href{http://bit.ly/1hGHVm8}{bit.ly/1hGHVm8}, retrieved 2014-08-19.} knowledge in the SP system is transparent and open to inspection, and likewise for the processing of knowledge.

\item {\em SP-neural}. As outlined in Section \ref{sp-neural_section}, the SP theory includes proposals---SP-neural---for how abstract concepts in the theory may be realised in terms of neurons and neural processes. The SP-neural proposals are significantly different from artificial neural networks as commonly conceived in computer science, and arguably more plausible in terms of neuroscience.

\end{itemize}

\section{Computational Efficiency, the Use of Energy, and the Bulkiness of Computers}\label{efficiency_energy_bulk_section}

With today's computers and related technologies, it would be difficult or impossible to make autonomous robots with anything approaching human-like versatility and adaptability:

\begin{quote}

``The human brain is a marvel. A mere 20 watts of energy are required to power the 22 billion neurons in a brain that's roughly the size of a grapefruit.\footnote{Another estimate is that, normally, the brain's power consumption is about 12.6 watts (``Does thinking really hard burn more calories?'', {\em Scientific American}, 2012-07-18, \href{http://bit.ly/1qJmCBG}{bit.ly/1qJmCBG}), and there may be as many as 86 billion neurons in the human brain \cite{herculano-houzel_2012}.} To field a conventional computer with comparable cognitive capacity would require gigawatts of electricity and a machine the size of a football field. ... Unless we can make computers many orders of magnitude more energy efficient, we're not going to be able to use them extensively as our intelligent assistants.'' \cite[p.~75, p.~88]{kelly_hamm_2013}.

\end{quote}

\noindent With AI and computer technologies as they are now, any computer that may approach human levels of intelligence would be too big to be mobile, and likewise for its power supply. And it seems that Moore's Law is unlikely to solve the problem of computational power, since that decades-long growth in the power of computer processors may tail off around 2020.\footnote{See, for example, ``Intel’s former chief architect: Moore's law will be dead within a decade'', {\em ExtremeTech}, 2013-08-30, \href{http://bit.ly/1niX9iK}{bit.ly/1niX9iK}. But for a contrary view, see ``Carbon nanotubes could step in to sustain Moore's Law'', {\em MIT Technology Review}, 117 (5), 17, 2014, \href{http://bit.ly/1nd49tD}{bit.ly/1nd49tD}.}

No doubt, there are many gains in efficiency that can be made with the von Neumann model of computing. But something radically different is likely to be needed to ``make computers many orders of magnitude more energy efficient'' and to achieve human-like intelligence with a computational device (with its power supply) that is portable.

The next three subsections describe how the SP system may help to achieve what is needed, firstly via compression of information, secondly via the exploitation of statistical information with heuristic techniques, and thirdly via a computer architecture that is inspired in part by Hebb's \cite{hebb_1949} concept of a ``cell assembly'', as outlined in Section \ref{sp-neural_section}. These proposals are an updated version of what's in \cite[Section IX]{sp_big_data}, with a shift of emphasis in Section \ref{data-centric_computing_in_sp-neural_section}.\footnote{In the SP computer model as it stands now, the use of heuristic techniques (Section \ref{heuristic_search_section}) is chiefly what allows the model to produce useful results with problems that would otherwise be intractable. But there appears to be considerable untapped potential for improvements in efficiency via the three approaches mentioned in the text.}

Apart from the concept of ``data-centric'' computing (referenced in Section \ref{data-centric_computing_in_sp-neural_section} and described in, for example, \cite{kelly_hamm_2013}), the proposals here, which all flow from the SP theory, are new contributions to thinking about the design of brains for autonomous robots.

\subsection{Efficiency Via Compression of Information}\label{efficiency_via_compression_of_information_section}

At the heart of the SP system is a process of searching for patterns that match each other. {\em Anything that increases the efficiency of searching is likely to increase the efficiency of computation.}

One thing that can increase the efficiency of searching is to reduce the size of the information to be searched. Here, the SP system may score because compression of information is central in how it works. Its repository of Old information is likely to be considerably smaller than the New information from which it was derived, so it should be correspondingly easy to search.

The SP system may also yield gains in efficiency via its system for creating relatively short codes for larger structures. If a short code can be used as a search term instead of the larger structure that it represents, searching is likely to be more efficient.

\subsection{Efficiency Via the Exploitation of Statistical Information, With Heuristic Techniques}\label{efficiency_via_statistical_information_section}

Continuing with the theme that anything that increases the efficiency of searching is likely to increase the efficiency of computation:

\begin{quote}

``If we want to find some strawberry jam, our search is more likely to be successful in a supermarket than it would be in a hardware shop or a car-sales showroom.'' \cite[section IX-A.2]{sp_big_data}.

\end{quote}

\noindent This seemingly-trite observation captures the essentials of the present proposal: instead of searching blindly for patterns that match each other, the system may use statistical knowledge, with heuristic techniques (Section \ref{heuristic_search_section}), to improve efficiencies.

In this efficiency-via-statistics concept, there is no need for any special process to gather statistical information. That information is collected because information compression is central in how the SP system works, and because of the intimate relation that exists between information compression and concepts of prediction and probability \cite{li_vitanyi_2009}. Those two things mean that the SP system, in the course of its operation, builds a statistical model of its data.

To flesh out this concept, it would probably be useful to examine it in quantitative terms, perhaps using the high-parallel version of the SP machine mentioned in Section \ref{outline_of_sp_system}. But in any study of that kind, it would be important to bear in mind that the effect of the concept in terms of computational efficiency is likely to depend on the computer architecture---real or simulated---in which it is applied. And to reap the full benefit of the concept, it will probably be necessary to apply it with an architecture that is quite different from the von Neumann model. One such possibility is outlined in the next subsection.

\subsection{Efficiency Via Pattern Assemblies}\label{efficiency_via_pattern_assemblies_section}

It is unlikely that problems of efficiency will be fully solved with robot brains in the von Neumann style. As argued in \cite[p.~9]{kelly_hamm_2013}, ``What's needed is a new architecture for computing, one that takes more inspiration from the human brain.'' The SP system has potential as a foundation for that new architecture.

SP-neural, the neural-inspired version of the SP theory outlined in Section \ref{sp-neural_section}, may help to promote computational efficiency in three main ways, as described in the next three subsections.

\subsubsection{Exploiting Statistical Information in SP-Neural}\label{statistical_information_in_sp-neural_section}

SP-neural suggests one possible means of realising the idea, outlined in Section \ref{efficiency_via_statistical_information_section}, that computational efficiency may be enhanced by taking advantage of the statistical information that the SP system gathers as an integral part of how it works.

It envisaged that, associated with each pattern assembly, will be some physiological analogue of the frequency of occurrence that is associated with the corresponding SP pattern (Section \ref{ic_prediction_probabilities_section}, \cite[Section 11.3.5]{wolff_2006}). We may suppose that, in the course of building neural analogues of multiple alignments, those physiological measures of frequency may serve to derive physiological analogues of the probabilities associated with multiple alignments, so that `good' multiple alignments may be distinguished from `bad' ones.

With that kind of mechanism, processing in SP-neural would be guided by statistical information as suggested in Section \ref{efficiency_via_statistical_information_section}. And since low-probability structures would be continually eliminated, we may suppose that much unnecessary processing would be avoided, with corresponding gains in computational efficiency.

\subsubsection{Cutting Out Searching in SP-Neural}\label{cutting_out_searching_in_sp-neural_section}

A potentially-important feature of SP-neural is that the hard-wired connections between pattern assemblies are always between symbols that match each other---and that can cut out a lot of searching. For example, the direct connection that can be seen in Figure \ref{connections_figure} between `\texttt{D}' in `\texttt{< D 0 t h i s >}' and `\texttt{D}' in `\texttt{< NP < D > < N > >}' means that there is no need for any kind of searching to find the match between those two instances of `\texttt{D}'.

Since, as was noted in Section \ref{efficiency_via_statistical_information_section}, ``anything that increases the efficiency of searching is likely to increase the efficiency of computation'', these hard-wired connections in SP-neural should mean increases in computational efficiency. These gains in computational efficiency are potentially large.

\subsubsection{Data-Centric Computing in SP-Neural}\label{data-centric_computing_in_sp-neural_section}

In \cite[Section IX-B]{sp_big_data}, it is suggested that the SP system may promote computational efficiency via a synergy with ``data-centric'' computing, meaning computing in which ``data processing [is] distributed throughout the computing system rather than concentrated in a CPU.'' \cite[p.~9]{kelly_hamm_2013} and ``the processing and the memory [is] closely integrated to reduce the shuttling of data and instructions back and forth.'' ({\em ibid.}).

That suggestion is not exactly wrong since, in the SP system, there is indeed a close integration of data and processing. But it needs to be qualified by two observations: 1) that there is really no place in the SP system for the concept of ``instruction''; and 2) that the process of finding matches between different portions of data may require transfers of information over relatively long distances. Regarding that last point, we cannot avoid such transfers by retreating to conventional architectures. It appears to be an unavoidable aspect of any system that aspires to human-like intelligence.

\subsection{Making Computers Smaller and Lighter}\label{bulkiness_section}

If computational efficiencies can be increased, as outlined above, it seems likely that there could be corresponding reductions in the size and weight of computers. The development of SP-neural as a model for computing may also lead to reductions in the bulkiness of computers.

\section{Towards Human-Like Versatility in Intelligence}\label{towards_versatility_section}

The current generation of robots fall far short of human-like versatility in intelligence. If they have anything approaching human-like intelligence it is almost always in a narrow field such as driving a car\footnote{See, for example, ``Autonomous car'', {\em Wikipedia}, \href{http://bit.ly/QKn6dg}{bit.ly/QKn6dg}, retrieved 2014-10-28.} or playing pool \cite{greenspan_etal_2008}.

This section aims to demonstrate how the SP system may promote human-like versatility in autonomous robots in several areas. Unsupervised learning is the main focus of Section \ref{towards_adaptability_section}, while other aspects of intelligence are considered in Sections \ref{simplification_integration_section} to \ref{sequential_and_parallel_procedures_section} below.

Versatility in intelligence---a major strength of the SP system---flows from the goal that has been central in the development of the theory: to combine conceptual {\em simplicity} with descriptive and explanatory {\em power}. This strength of the SP system chimes well with what is required in any autonomous robot that is to function effectively in situations where little or no help can be provided by people.

\subsection{Simplification and Integration}\label{simplification_integration_section}

Before getting on to specifics, we shall consider simplification and integration, and their importance in the design of autonomous robots. In that connection, some may argue that human-like versatility could be achieved with a collection of applications, each one dedicated to a particular aspect of intelligence. That being so, the argument may run, there is no need for an all-in-one solution like the SP system. But human intelligence is not like that:

\begin{itemize}

\item Each of our concepts is, normally, a seamlessly-integrated amalgam of different kinds of knowledge. For example, most people's concept of a ``steam engine'' includes static and moving images, sounds and smells, associations with journeys by steam train, literary, historical and technical knowledge, and more.

\item There is smooth inter-working of different aspects of human intelligence---learning, recognition, reasoning, problem-solving, and so on---without artificial barriers or transitions.

\end{itemize}

By contrast, a collection of AI-related applications would be merely a kludge that is likely to suffer from poor integration of knowledge structures and awkward incompatibilities between different subsystems.

The key difference between the SP system and that kind of assemblage of AI-related applications is its central organising principle:  simplification and integration of concepts across artificial intelligence, mainstream computing, mathematics, and human perception and cognition. There is potential in that principle for autonomous robots to achieve the kind of seamless integration of diverse kinds of knowledge and diverse kinds of processing that is a hallmark of human intelligence.

The following three subsections expand on aspects of the principle and its apparent importance in the design of autonomous robots.

\subsubsection{Simplification of Structures and Functions}

Simplification in the SP system flows from two main things:

\begin{itemize}

\item The adoption of one simple format---SP patterns---for the representation of all kinds of knowledge.

\item One computational framework, with multiple alignment centre-stage, for all kinds of processing.

\end{itemize}

Like a database management system or a `shell' for an expert system, the SP system provides one framework that can be loaded with different bodies of knowledge according to need. As with database management systems (DBMSs) and expert-system shells, this cuts out the need to recreate the framework for each new application, meaning that there can be a substantial overall simplification across a range of applications \cite[Section 5]{sp_benefits_apps}. The main difference between the SP system and any DBMS or expert-system shell is the versatility of the multiple alignment framework, especially in AI-related functions.

In the design of autonomous robots, this kind of simplification in software may have some impact on the bulkiness of robot brains. But potentially more important is how simplification of software may simplifying the creation, operation, and management of software, with corresponding gains in efficiency. Even if the creation, operation, and management of software were to be fully automated---as is envisaged for the SP system---gains in efficiency are potentially significant.

\subsubsection{Integration of Structures and Functions}\label{integration_section}

Using one simple format for all kinds of knowledge and one computational framework for all kinds of processing is likely to yield a second benefit: the afore-mentioned seamless integration of diverse kinds of knowledge and smooth inter-working of different functions. Here are some putative examples:

\begin{itemize}

\item {\em Syntax and semantics}. It is clear that in the understanding of any natural language (listening or reading) and the production of language (speaking or writing), there must be close integration and inter-working of the syntactic forms of language and what they mean. Achieving that intimate relationship between syntax and semantics is likely to be made easier by using what the SP system provides: one simple format for both syntax and semantics and one computational framework for all kinds of knowledge. In support of that idea, it is known that, at least in English, some aspects of syntax cannot be defined except with reference to semantics \cite[Section 6.2]{sp_benefits_apps}.

\item {\em Recognition and learning}. Although recognition and learning may be treated as distinct topics in text books, they are difficult to separate in practice. Consider, for example, how a dog chases and catches a ball that is bouncing haphazardly across uneven ground. The dog must, of course, recognise the ball but, after every bounce, he or she must be constantly assimilating new information about the speed and the direction of travel of the ball. The SP system may facilitate that kind of close integration of recognition and learning by providing one simple format for knowledge and one computational framework for both recognition and learning.

\item {\em Knowledge representation and learning}. If lightning is represented in several different ways and likewise with thunder then, as noted in \cite[Section III-A.1]{sp_big_data}, it would be difficult or impossible to learn, without being told, that there is an association between those two things. For a learning system to detect the way in which lightning is normally followed by thunder, it seems necessary to get behind the variability of surface forms and derive new structures and associations from the underlying knowledge, expressed in some kind of universal framework for the representation and processing of diverse kinds of knowledge (UFK). Likewise for other structures and associations.

\item \sloppy {\em Knowledge representation and reasoning}. In any kind of criminal investigation---the subject of countless television dramas---there must be total flexibility to use any kind of knowledge---physical, chemical, social, psychological, legal, and so on---to narrow the field of suspects and find ``whodunnit''. It is difficult to see how that flexibility may be achieved except by the use of a uniform format for all kinds of knowledge and one computational framework for the processing of those diverse kinds of knowledge.

\end{itemize}

In general, the SP system provides for seamless integration of structures and functions, in any combination, in all the areas described in Sections \ref{natural_language_processing_section} to \ref{sequential_and_parallel_procedures_section}.

\subsubsection{Deeper Insights and Better Solutions to Problems}

The quest for simplification and integration in the SP theory accords with Occam's Razor, one of the most widely-accepted principles in science. In terms of that principle, the SP theory scores well, since a relatively simple framework provides an account of a wide range of concepts and phenomena.

An often-repeated observation in science is that a good theory can yield `deeper' insights and better solutions to problems than would otherwise be possible. For example, Einstein's theory of general relativity led to the prediction---confirmed by observation---that light would be bent by gravity, and it provided an account of the precession of the perihelion of Mercury.

The SP theory is beginning to show benefits of that kind:

\begin{itemize}

\item Relatively new insights are the ways in which computational efficiency may be improved, with corresponding savings in the use of energy, via aspects of the SP system described in Section \ref{efficiency_energy_bulk_section}.

\item Other insights, summarised in \cite[Section 6]{sp_benefits_apps}, including: a new and apparently improved method for encoding discontinuous dependencies in syntax; seamless integration of class-inclusion hierarchies and part-whole hierarchies; the use of one framework for both the analysis and production of knowledge; and benefits for relational, object-oriented, and network models for databases.

\end{itemize}

\subsection{Natural Language Processing}\label{natural_language_processing_section}

This and the following subsections, together with Section \ref{towards_adaptability_section} (human-like adaptability), demonstrate some of the versatility of the SP system in areas that are likely to prove useful in autonomous robots.

In addition to the learning of linguistic knowledge (Section \ref{learning_linguistic_knowledge_section}), the SP system has strengths in the parsing of natural language, the production of natural language, and the integration of syntactic and semantic knowledge, as outlined in this section. These aspects of the system are described more fully in \cite[Section 8]{sp_extended_overview} and in \cite[Chapter 5]{wolff_2006}.

\subsubsection{Parsing of Natural Language}\label{parsing_nl_section}

Figure \ref{parsing_figure} shows how, via multiple alignment, a sentence (in row 0) may be parsed in terms of grammatical structures including words (rows 1 to 8).\footnote{Compared with the multiple alignment shown in Figure \ref{class_part_plant_figure}, this multiple alignment is rotated through $90\degree$, replacing columns with rows. The choice between these two styles, which are equivalent, depends largely on what fits best on the page.} It also shows, in row 8, how the system may mark the syntactic dependency between the plural subject of the sentence (`\texttt{Np}') and the plural main verb (`\texttt{Vp}') (see also \cite[Sections 5.4 and 5.5]{wolff_2006}, \cite[Section 8.1]{sp_extended_overview}).

\begin{figure*}[!htbp]
\fontsize{07.00pt}{08.40pt}
\centering
{\bf
\begin{BVerbatim}
0                       t h e                a p p l e    s                a r e         s w e e t       0
                        | | |                | | | | |    |                | | |         | | | | |
1                       | | |         N Nr 6 a p p l e #N |                | | |         | | | | |       1
                        | | |         | |              |  |                | | |         | | | | |
2                       | | |    N Np N Nr             #N s #N             | | |         | | | | |       2
                        | | |    | |                        |              | | |         | | | | |
3                  D 17 t h e #D | |                        |              | | |         | | | | |       3
                   |          |  | |                        |              | | |         | | | | |
4            NP 0a D          #D N |                        #N #NP         | | |         | | | | |       4
             |                     |                            |          | | |         | | | | |
5            |                     |                            |  V Vp 11 a r e #V      | | | | |       5
             |                     |                            |  | |           |       | | | | |
6 S Num    ; NP                    |                           #NP V |           #V A    | | | | | #A #S 6
     |     |                       |                                 |              |    | | | | | |
7    |     |                       |                                 |              A 21 s w e e t #A    7
     |     |                       |                                 |
8   Num PL ;                       Np                                Vp                                  8
\end{BVerbatim}
}
\caption{The best multiple alignment created by the SP model with a store of Old patterns like those in rows 1 to 8 (representing grammatical structures, including words) and a New pattern (representing a sentence to be parsed) shown in row 0. Reproduced from Figure 1 in \protect\cite{sp_benefits_apps}, with permission.}
\label{parsing_figure}
\end{figure*}

To create a multiple alignment like the one in the figure, the system needs a grammar of Old patterns, like those shown, one per row, in rows 1 to 8 of the figure. In this example, the patterns represent linguistic structures including words.

Although SP patterns are remarkably simple, it appears that, within the multiple alignment framework, they have at least the expressive power of a context-sensitive grammar \cite[Sections 5.4 and 5.5]{wolff_2006}. As previously noted (Section \ref{patterns_and_symbols_section}), there is reason to believe that all kinds of knowledge may be represented, within the multiple alignment framework, by SP patterns.

\subsubsection{Production of Natural Language}

A neat feature of the SP system is that one set of mechanisms and processes may achieve both the analysis or parsing of natural language (Section \ref{parsing_nl_section}) and the generation or production of sentences. This is explained in \cite[Section 3.8]{wolff_2006} and \cite[Section 4.5]{sp_extended_overview}.

\subsubsection{The Integration of Syntax and Semantics}\label{syntax_semantics_section}

The use of one simple format for all kinds of knowledge is likely to facilitate the seamless integration of syntax and semantics. Preliminary examples of how this may be done are shown in \cite[Section 5.7]{wolff_2006}, both for the derivation of meanings from surface forms \cite[Figure 5.18]{wolff_2006} and for the production of surface forms from meanings \cite[Figure 5.19]{wolff_2006}.

\subsubsection{Parallel Streams of Information}\label{nl_parallel_streams_section}

Up to now, most work on natural language within the SP research programme has made the simplifying assumption that natural language may be represented with a sequence of symbols, as in ordinary text. But this 1D assumption does not sit easily with some aspects of natural language:

\begin{itemize}

\item Vowel sounds, for example, may be analysed into formants, two or more of which may occur simultaneously. Vowels, and perhaps other elements of speech, may be represented most naturally with parallel streams of information.

\item It does not seem right that the syntactic and semantic aspects of natural language should be forced into the procrustean bed of a single sequence. As with formants in speech, it seems most natural to regard syntax and semantics as parallel streams of information.

\end{itemize}

The way in which parallel streams of information may be represented and processed in the SP system is described in Appendix \ref{parallel_streams_of_information_section}.

\subsection{Pattern Recognition}\label{pattern_recognition_section}

As described quite fully in \cite[Chapter 6]{wolff_2006} and more briefly in \cite[Section 9]{sp_extended_overview}, the SP system has strengths in several aspects of pattern recognition:

\begin{itemize}

\item It can recognise patterns at multiple levels of abstraction, with the integration of class-inclusion relations and part-whole relations, as shown in the example in Figure \ref{class_part_plant_figure}.

\item It can model ``family resemblance'' or polythetic categories, meaning that recognition does not depend on the presence absence of any particular feature or combination of features.

\item Recognition is robust in the face of errors of omission, commission or substitution in the New pattern or patterns.

\item For any given identification, or any related inference, the SP system may calculate associated probabilities.

\item As a by-product of how recognition is achieved via the building of multiple alignments, the system provides a model for the way in which context may influence recognition.

\end{itemize}

The SP system also provides a framework for medical diagnosis via pattern recognition, with potential for diagnosis via causal reasoning \cite{wolff_medical_diagnosis}, \cite[Section 6.5]{wolff_2006}.

\subsection{Information Storage and Retrieval}\label{storage_retrieval_section}

The SP system may serve as a database system with `intelligence' \cite{wolff_sp_intelligent_database}, \cite[Chapter 6]{wolff_2006}. Although this may seem somewhat removed from the world of autonomous robots, any such robot will need such `database' functions as storing information and retrieving it. Apart from aspects of intelligence (as outlined elsewhere in this paper), the main strengths of the SP system are:

\begin{itemize}

\item The system lends itself to information retrieval in the manner of query-by-example. There is also potential for information retrieval via the use of natural language or query languages such as SQL.

\item As outlined in Section \ref{ma_knowledge_section}, the system supports object-oriented concepts such as class hierarchies (including cross-classification), and inheritance of attributes, and it provides for the representation of part-whole hierarchies and their seamless integration with class hierarchies. The system also supports network, relational, and entity-relationship database models.

\end{itemize}

\subsection{Vision}\label{vision_section}

With generalisation of the SP system to accommodate 2D patterns, it has potential to model several aspects of natural vision and to facilitate the development of human-like abilities in robot vision \cite{sp_vision}. In these connections, the main strengths and potential of the SP system are:

\begin{itemize}

\item Low level perceptual features such as edges or corners may be identified via the multiple alignment framework by the extraction of redundancy in uniform areas in the manner of the run-length encoding technique for information compression \cite[Section 3]{sp_vision}.

\item The system may be applied in the recognition of objects and in scene analysis, with the same strengths as in pattern recognition (Section \ref{pattern_recognition_section}).

\item There is potential for the learning of visual entities and classes of entity (Section \ref{learning_to_see_section}) and the piecing together of coherent concepts from fragments \cite[Section 5.4]{sp_vision}.

\item There is potential, via multiple alignment, for the creation of 3D models of objects and of a robot's surroundings (Section \ref{building_model_of_world_section}).

\item The SP theory provides an account of how we may see things that are not objectively present in an image, how we may recognise something despite variations in the size of its retinal image, and how raster graphics and vector graphics may be unified.

\item And the SP theory has things to say about the phenomena of lightness constancy and colour constancy, ambiguities in visual perception, and the integration of vision with other senses and other aspects of intelligence.

\end{itemize}

\subsection{Reasoning}\label{reasoning_section}

As described in quite fully in \cite[Chapters 7 and 10, Section 6.4]{wolff_2006} and more selectively in \cite[Section 10]{sp_extended_overview}, the SP system lends itself to several kinds of reasoning:

\begin{itemize}

\item One-step `deductive' reasoning.

\item Abductive reasoning.

\item Reasoning with probabilistic decision networks and decision trees.

\item Reasoning with `rules'.

\item Nonmonotonic reasoning and reasoning with default values.

\item Reasoning in Bayesian networks, including ``explaining away''.

\item Causal diagnosis.

\item Reasoning which is not supported by evidence.

\item Inheritance of attributes in an object-oriented class hierarchy or heterarchy.

\end{itemize}

In keeping with the remarks about integration in Section \ref{integration_section}, these several kinds of reasoning may work together seamlessly without awkward incompatibilities, and likewise for how they may integrate seamlessly with such AI functions as unsupervised learning, pattern recognition, and so on.

For any given inference reached via any of these kinds of reasoning, the SP system may calculate associated probabilities (Section \ref{ic_prediction_probabilities_section}). Although the system is fundamentally probabilistic, it may imitate the effect of logic and other `exact' forms of reasoning \cite[10.4.5]{wolff_2006}.

\subsubsection{Spatial Reasoning}\label{spatial_reasoning_section}

If, as seems likely, multiple alignment provides a means for an autonomous robot to build a 3D model of objects and of its surroundings (Section \ref{building_model_of_world_section}), this may open the door to spatial reasoning. There is potential, for example, for a robot to explore `mentally' how furniture may be arranged in a room, much as people sometimes use cardboard shapes representing furniture, with a plan of a room, to work out how things may be fitted together.

\subsubsection{What-If Reasoning}

Although a flight simulator is not normally regarded as a system for reasoning, it provides a very effective means of exploring what may happen if, for example, a plane loses power in one of its engines or if there is ice on the wings.

Similar things may apply with knowledgeable robots. In view of the versatility of the SP system in processing knowledge of various kinds (Section \ref{ma_knowledge_section}), and in view of the system's capabilities and potential in reasoning, mentioned above, there is potential for the system to explore what-if scenarios arising from this or that hypothetical contingency.

\subsection{Planning and Problem Solving}

With data about flights between different cities, represented using SP patterns, the SP computer model may find a route between any two cities (if such a route exists) and, if there are alternative routes, it may find them as well \cite[Section 8.2]{wolff_2006}.

Provided they are translated into textual form, the SP70 computer model can solve geometric analogy problems of the kind found in puzzle books and some IQ tests \cite[Section 8.3]{wolff_2006}, \cite[Section 12]{sp_extended_overview}.

\subsection{Sequential and Parallel Procedures}\label{sequential_and_parallel_procedures_section}

Although it may not seem obvious at first sight, the multiple alignment framework can model several devices used in ordinary procedural programming, including: {\em procedure}, {\em function}, or {\em subroutine}; {\em variable}, {\em value} and {\em type}; {\em function with parameters}; {\em conditional statement}; and the means of repeating operations such as {\em repeat ...~until} or {\em do ...~while} \cite[Section 6.6.1]{sp_benefits_apps}. In accordance with good practice in software engineering, the SP system facilitates the integration of `programs' with `data'. And as previously noted (Section \ref{ma_knowledge_section}), the SP system supports object-oriented concepts such as class hierarchies with inheritance of attributes.

In \cite[Section 6.6.3]{sp_benefits_apps}, it is suggested that, since SP patterns at the `top' level are independent of each other, they may serve to model processes that may run in parallel. Now it appears that a better option is to model parallel processes as parallel streams of information, represented in 2D patterns, as described in Appendix \ref{parallel_streams_of_information_section}. The advantage of this latter scheme is that it provides the means of showing when two or more events occur at the same time, and the relative timings of events.

Within the SP system, these structures and mechanisms may serve in the representation and processing of sequential and parallel procedures from the real world such as those required for cooking a meal, organising a party, going shopping, and so on.

\section{Towards Human-Like Adaptability in Intelligence}\label{towards_adaptability_section}

As with versatility in intelligence (Section \ref{towards_versatility_section}), the current generation of robots falls far short of human-like adaptability in intelligence. The key to that adaptability is the ability to learn, an aspect of human-like versatility in intelligence (Section \ref{towards_versatility_section}) but considered here in a separate section because of its importance.

As with efficiency in computations and versatility in intelligence, the SP system promises solutions for learning and adaptability that are rather different from others in the field of robotics and with considerable potential for autonomous robots.

After a `preliminaries' subsection, the main elements of learning in the SP system are described. Subsections that follow describe several aspects of how the SP theory may be applied to learning in autonomous robots. These parts of the paper develop important refinements in the SP theory that are needed to meet the demands of this area of application.

\subsection{Preliminaries}\label{preliminaries_section}

Here we consider, first, some forms of learning and then how rewards and punishments (carrots and sticks) may relate to learning. Two subsections that follow outline the research on the learning of a first language that provided part of the inspiration for the SP theory, and the reorganisation that has been needed to meet the goals of the SP research programme.

\subsubsection{Forms of Learning}\label{forms_of_learning_section}

In the preceding parts of this paper, the word ``learning'' has generally been preceded by the word ``unsupervised''. That qualification means that learning occurs without the benefit of any kind of ``teacher'', or the grading of learning materials from simple to complex, or the provision of ``negative'' examples of concepts to be learned, meaning examples that are marked as ``wrong'' ({\em cf.} \cite{gold_1967})---and it also means that learning occurs without rewards or punishments (Section \ref{carrots_sticks_section}). Learning with assistance from those kinds of things is, of course, ``supervised'' learning.

We may also distinguish between {\em primary} forms of learning---the learning of basic skills---and {\em secondary} forms of learning---the kinds of learning that depend on those basic skills. As an example, learning via lessons in schools and colleges may be regarded as a secondary form of learning that depends on such basic skills as being able to speak and to understand speech. Learning by watching and imitating what other people do may also be regarded as a secondary form of learning because it depends on more basic skills such as the ability to interpret visual inputs in terms of people and the motions of their limbs, hands, feet, and so on.

Of course, there is really a hierarchy of skills and corresponding forms of learning, because a skill such as the ability to read---which would normally be learned as a secondary skill in school---may itself provide a basis for learning from newspapers, magazines, and books, from the internet, and so on.

This paper focusses mainly on the learning of foundation skills. And a working hypothesis---with supporting evidence from research on the learning of a first language or languages \cite{wolff_1988}---is that unsupervised learning is an important driver in both primary and secondary forms of learning.

Achieving a good theory of unsupervised learning may, arguably, be seen as the ``Holy Grail'' of research on learning, especially learning by autonomous robots. Notwithstanding the undoubted importance of schools and colleges, it appears that much of what we know is picked up via our everyday experiences, without explicit teaching. Unsupervised learning is what an autonomous robot would need in places like Mars where it can get little or no help from a human teacher. And if we can develop a good theory of unsupervised learning, it should smooth the path in understanding secondary forms of learning.

\subsubsection{Carrots, Sticks, and Motivations}\label{carrots_sticks_section}

In research on human and animal learning, most famously the work of psychologist B.~F.~Skinner,\footnote{See, for example, ``B.~F.~Skinner'', {\em Wikipedia}, \href{http://bit.ly/X5BJuH}{bit.ly/X5BJuH}, retrieved 2014-09-15.} there is a long tradition of linking learning with `reinforcements', both positive (rewards or `carrots') and negative (punishments or `sticks'). This relates to Ivan Pavlov's research on classical conditioning,\footnote{See, for example, ``Ivan Pavlov'', {\em Wikipedia}, \href{http://bit.ly/1mOEmuY}{bit.ly/1mOEmuY}, retrieved 2014-09-15.} and the long-established practice by animal trainers of using small rewards to encourage some kinds of behaviour and mild punishments to discourage others.

It is clear from the research, and from successes in the training of animals, that carrots and sticks can be very effective. Motivation is certainly relevant to learning. But in the development of the SP theory, no attempt has been made to say anything about reinforcements, or about related concepts such as motivation. Here are some tentative thoughts:

\begin{itemize}

\item In view of experimental evidence that learning can occur without reinforcement,\footnote{See, for example, ``Latent learning'', {\em Wikipedia}, \href{http://bit.ly/1pluRo3}{bit.ly/1pluRo3}, retrieved 2014-08-16.} it seems unlikely that the concept of reinforcement could be central in any comprehensive theory of learning.

\item It seems that concepts of reinforcement are unlikely to take us very far in explaining the learning of complex forms of knowledge and behaviour such as natural languages. In that connection, Chomsky's critique of Skinner's book on verbal behaviour and how it may be learned \cite{chomsky_1959} is still worth reading, despite the passage of time.

\item In animal training, it seems that reinforcement is chiefly a means of communicating to the animal what the trainer would like it to do. In that perspective, reinforcement may help to create new combinations of pre-existing forms of behaviour but would have little or no role in the creation of those pre-existing behaviours.

\item An association between a reward (or a punishment) and a particular form of behaviour is just one of many types of redundancy that a person or robot needs to learn. Detecting or discovering that kind of association may be achieved within the SP system via the same mechanisms and processes that serve in the discovery of other kinds of redundancy in information.

\item It appears that the principle of minimum length encoding can provide a foundation not only for grammatical inference but more generally for unsupervised learning, for other aspects of perception and cognition, and for secondary forms of learning mentioned in Section \ref{forms_of_learning_section}.

\item Notwithstanding the foregoing points, there is a clear need to expand our understanding of the relationship between learning and concepts of reinforcement and motivation. Why do children play? (Section \ref{exploration_play_section}); What induces people to spend thousands of hours learning the skills needed for pool, billiards, or snooker? (Section \ref{learning_skills_section}); And so on.

\end{itemize}

\subsubsection{Computer Models of Language Learning}\label{computer_models_of_language_learning_section}

Part of the inspiration for the SP theory has been an earlier programme of research on grammatical inference, developing computer models of the unsupervised learning of a first language or languages. There is an overview of this research in \cite{wolff_1988}. The main conclusions are:

\begin{itemize}

\item That the principle of minimum length encoding is of key importance in understanding the unsupervised learning of a first language. In brief, this means that, in inferring a grammar from a body of data, {\bf I}, we should aim to minimise ($G + E$), where $G$ is the size of the grammar, {\bf G}, derived from {\bf I}, and $E$ is the size of the encoding of {\bf I} in terms of {\bf G} (an encoding that may be referred to as {\bf E}).

\item The ``MK10'' model, based on the MLE principle, can successfully discover word structure in an unsegmented sample of natural language without any kind of dictionary, except what it creates for itself. With a little human assistance, the MK10 model can discover phrase structure as well.

\item The more fully-developed ``SNPR'' model, based on the MLE principle, can derive a plausible generative grammar from an unsegmented sample of English-like artificial language, including segmental structures, classes of structure, and abstract patterns.

\item The principle of minimum length encoding provides an explanation for aspects of language learning that may otherwise be puzzling:

\begin{itemize}

\item {\em Generalisation from a finite sample}. How we can generalise from the finite (albeit large) sample of a given language which is the basis for our learning to a knowledge of the language that embraces an infinite range of utterances,\footnote{That a natural language like English embraces an infinite range of utterances can be seen from recursive structures like {\em This is the cow with the crumpled horn That tossed the dog that worried the cat That chased the rat that ate the cheese ...}. Native speakers know that there is, in principle, no limit on the length of such sentences, or their variety.}

\item {\em How over-generalisations may be corrected}. It is well known that young children do over-generalise grammatical rules---for example, they may say ``buyed'' as the past tense of ``buy'' instead of ``bought''---but those kinds of over-generalisation disappear later.

    A possible explanation is that children's errors are corrected by adults, but there is evidence that the learning of a first language does not depend on the correction of errors by adults, or anything equivalent. The principle evidence is that children with a physical handicap that prevents them speaking intelligibly may, nevertheless, learn to comprehend their native language successfully \cite{lenneberg_1962,brown_1989}. Since they say little or nothing that adults can understand, there is little that adults can do to correct any errors.

\item {\em Learning from dirty data}. People can distinguish sharply between utterances that belong in their native language and those that don't, and this despite the fact that, normally, much of the language that we hear as children is not grammatically correct. As before, there is evidence that children can learn their native language without the benefit of error-correction by adults or anything equivalent \cite{lenneberg_1962,brown_1989}.

\end{itemize}

In summary, grammars that minimise ($G + E$) are ones that generalise without over-generalising and that filter out haphazard errors. Systematic `errors' (ones that are not haphazard) are likely to acquire the status of dialect forms and thus lose their status as errors.

\end{itemize}

There is more detail about these points in \cite[Section 5.3]{sp_extended_overview}. Incidentally, the Chomskian argument that children learn via an inborn knowledge of ``universal grammar'' does not bear scrutiny because it depends on the still unproven idea that there is substantial structure that is shared by all the world's languages and it is vague about how a child might learn the specifics of his or her native language.

As we shall see in Section \ref{generalisation_section}, these ideas appear to have some useful things to say about how learning can or should be developed in autonomous robots.

\subsubsection{Reorganisation Needed to Meet the Goals of the SP Research Programme}\label{reorganisation_in_sp_section}

In their broad organisation, the MK10 and SNPR computer models may be roughly characterised as hierarchical chunking models, building a knowledge of recurrent `chunks' of information via hierarchies of smaller elements. But although this organisation was quite successful in modelling aspects of language learning, it proved to be quite unsuitable for the goals of the SP research programme: simplification and integration of concepts across artificial intelligence, mainstream computing, mathematics, and human perception and cognition.

The multiple alignment framework that has been developed to meet those goals is quite different from the MK10 and SNPR models, but the principle of minimum length encoding remains centre-stage. The multiple alignment framework is much more general: it can model hierarchical chunking where that is appropriate but it can also model several other kinds of structure as well (Section \ref{ma_knowledge_section}).

Given that the SP research grew out of earlier research on language learning, it is pertinent to ask how the SP computer model performs in that area of application? In brief, the answer is ``much the same as earlier models but still not as well as one might wish''. More specifically, the SP computer model, like the SNPR model, can derive plausible generative grammars from samples of English-like artificial languages, including segmental structures, classes of structure, and abstract patterns \cite[Chapter 9]{wolff_2006}.

At present, the main shortcoming of the SP computer model with respect to learning is that it does not detect intermediate levels of structure and it cannot learn discontinuous dependencies in natural language syntax \cite[Section 3.3]{sp_extended_overview}. But, as previously noted ({\em ibid.}), I believe these problems are soluble and that solving them will greatly enhance the capabilities of the system for the unsupervised learning of structure in data. Apart from these developments, the SP computer model also needs to be generalised to work with patterns in two dimensions ({\em ibid.}).

Although, in the learning of syntactic structures, the SP computer model does much the same as earlier models, it has two decisive advantages: the integration of learning with other aspects of intelligence; and the potential to learn any of the kinds of knowledge described in Sections \ref{natural_language_processing_section} to \ref{sequential_and_parallel_procedures_section}, not merely syntactic kinds of knowledge.

\subsection{Unsupervised Learning in the SP System}\label{us_learning_in_sp_system_section}

Unsupervised learning by SP70 is described in outline in \cite[Section 5]{sp_extended_overview} and \cite[Section 3.9.2]{wolff_2006}, and in more detail in \cite[Section 9.2]{wolff_2006}. Here, the main features of the learning process are described as a basis for the proposals, in Sections \ref{learning_linguistic_knowledge_section} to \ref{cutting_learning_costs_section}, about how the same kinds of learning processes may be applied in autonomous robots.

In broad terms, the SP70 model processes a set of New patterns (which may be referred to as {\bf I}) in two main phases:

\begin{enumerate}

\item Create a set of Old patterns that may be used to encode {\bf I}.

\item From the Old patterns created in the first phase, compile one or more alternative grammars for the patterns in New, in accordance with principles of minimum length encoding.

\end{enumerate}

As noted in Section \ref{heuristic_search_section}, the process of inferring one or more good grammars from a body of data is normally too complex to be solved by exhaustive search. But, via the use of heuristic techniques (as outlined in Section \ref{compiling_alternative_grammars_section}, below), it is normally possible to find reasonably good solutions without undue computational demands.

Although the two phases just described have the flavour of batch processing, they may be adapted for an incremental style of working: processing New information as it is received and building collections of Old patterns that may serve in the economical encoding of New patterns that are received later.

The two phases are described in a little more detail in the following to subsections.

\subsubsection{Creating Candidate Patterns}\label{creating_candidate_patterns_section}

In SP70, candidate patterns for inclusion in the repository of Old patterns are derived from multiple alignments like the one shown in Figure  \ref{learning_alignment_figure}. Here, the pattern shown in row 1 is an analogue of something that a child has heard (`\texttt{t h a t b o y r u n s}') with the addition of code symbols `\texttt{<}', `\texttt{\%1}', `\texttt{9}', and `\texttt{>}', while the pattern in row 0 (`\texttt{t h a t g i r l r u n s}') is an analogue of something that the same child has heard later. The letters are analogues of symbols in speech (see Appendix \ref{identification_of_symbols_section}).

\begin{figure}[!htbp]
\centering
{\bf
\begin{BVerbatim}
0        t h a t g i r l r u n s   0
         | | | |         | | | |
1 < %1 9 t h a t b o y   r u n s > 1
\end{BVerbatim}
}
\caption{A simple multiple alignment from which patterns may be derived. Reproduced from Figure 9.2 in \cite{wolff_2006}, with permission.}
\label{learning_alignment_figure}
\end{figure}

From that multiple alignment, the program derives the patterns `\texttt{t h a t}' and `\texttt{r u n s}' from subsequences that match each other, and it derives `\texttt{g i r l}' and `\texttt{b o y}' from subsequences that don't match. In addition, the program assigns code symbols to the newly-created patterns so that `\texttt{t h a t}' becomes `\texttt{< \%7 12 t h a t >}', `\texttt{r u n s}' becomes `\texttt{< \%8 13 r u n s >}', and so on. And, using those code symbols, the program creates an abstract pattern, `\texttt{< \%10 16 < \%7 > < \%9 > < \%8 > >}', that records the whole sequence.

The overall result in this example is the set of patterns shown in Figure \ref{learning_patterns_figure}. This is essentially a simple grammar for sequences of the form `\texttt{t h a t g i r l r u n s}' and `\texttt{t h a t b o y r u n s}'.

\begin{figure}[!htbp]
\centering
{\bf
\begin{BVerbatim}
< %7 12 t h a t >
< %9 14 b o y >
< %9 15 g i r l >
< %8 13 r u n s >
< %10 16 < %7 > < %9 > < %8 > >
\end{BVerbatim}
}
\caption{Patterns derived from the multiple alignment shown in Figure \ref{learning_alignment_figure}. Reproduced from Figure 9.3 in \cite{wolff_2006}, with permission.}
\label{learning_patterns_figure}
\end{figure}

\subsubsection{Compiling Alternative Grammars}\label{compiling_alternative_grammars_section}

The example just described shows how SP70 creates candidate patterns and grammars via partial matching but the tidiness of the multiple alignment in Figure \ref{learning_alignment_figure} and of the grammar shown in Figure \ref{learning_patterns_figure} may be misleading. In practice, the system creates many multiple alignments that are less neat than the one shown and many candidate grammars that are intuitively `wrong'. But the wrong grammars are progressively weeded out, as described next.

As noted in Section \ref{heuristic_search_section}, the system exploits principles of heuristic search to cope with complexity: for any given body of New patterns, {\bf I}, the system explores the abstract space of possible grammars in stages; at each stage, in accordance with the principle of minimum length encoding (Section \ref{computer_models_of_language_learning_section}), each candidate grammar is evaluated in terms of the size of ($G + E$); and, at each stage, grammars that perform poorly are discarded. In this way, the system may gradually develop one or more {\bf G}s that perform reasonably well in terms of ($G + E$), and it may achieve these results without unreasonable demands on computational resources.

The patterns in a successful grammar are ones that express redundancy (repetition of information) in {\bf I}. As a rough generalisation, these are ones that occur frequently, or are relatively large in terms of the amount of {\bf I} that they may encode, or both these things. Exceptions to that rule are patterns that play supporting roles.

As a general rule---the DONSVIC principle described in \cite[Section 5.2]{sp_extended_overview}---grammars that minimise ($G + E$) are also ones that appear `natural' to people.

\subsection{One-Trial Learning}\label{one_trial_learning_section}

This subsection and the ones that follow discuss aspects of how the SP system may be applied to learning in autonomous robots.

Unsupervised learning in the SP theory \cite[Chapter 9]{wolff_2006}, \cite[Section 5]{sp_extended_overview} is quite different from ``Hebbian learning''---gradual strengthening of links between neurons---that is widely adopted in the kinds of artificial neural networks that are popular in computer science. By contrast with Hebbian learning, the SP system, like a person, may learn from a single exposure to some situation or event.\footnote{This is because the first step in unsupervised learning in the SP system is for the system to take in information via its senses, as indicated in Section \ref{creating_candidate_patterns_section}.} And, by contrast with Hebbian learning, it takes time to learn a language in the SP system (Section \ref{learning_linguistic_knowledge_section}), or to learn the skills needed for games like pool, billiards, or snooker (Section \ref{learning_skills_section}), because of the complexity of the search space, not because of any kind of progressive ``weighting'' of links between neurons \cite[Section 11.4.4]{wolff_2006}.

Donald Hebb \cite{hebb_1949} recognised that his central mechanism for learning could not account for one-trial learning and introduced a `reverberatory' mechanism to plug the gap. But this has its own problems as outlined in \cite[Section 11.4.4.1]{wolff_2006}.

\subsection{Learning Linguistic Knowledge}\label{learning_linguistic_knowledge_section}

It appears that the SP system has considerable potential as a framework for the learning of linguistic knowledge, including syntax, semantics, and their integration.

Semantic knowledge---the non-syntactic meanings of speech or writing---would include the kinds of knowledge discussed in Sections \ref{natural_language_processing_section} to \ref{sequential_and_parallel_procedures_section}. Aspects of how such knowledge may be learned are discussed in Sections \ref{learning_to_see_section} to \ref{generalisation_section}.

In learning how words relate to meanings, a child (or robot) must solve the problem that, when a word is heard in a given physical context, it may refer to any aspect of that context or to something else entirely.\footnote{{\em cf.} Quine's discussion of how a linguist that is studying an unfamiliar language might infer the meaning of ``Gavagai'' \cite[Chapter 2]{quine_1960}.} In the SP system, this problem may be solved statistically: in one context, a given word is likely to be ambiguous; but across several different contexts, associations are likely to emerge via the discovery of redundancies in the data.

As noted in Section \ref{nl_parallel_streams_section}, it seems most natural to regard syntax and semantics as parallel streams of information. How unsupervised learning may be applied with that kind of information is discussed in Section \ref{interactions_regularities_section}, drawing on ideas presented in Appendix \ref{parallel_streams_of_information_section}.

\subsection{Learning to See}\label{learning_to_see_section}

In the same way that the SP computer model may build grammars for one-dimensional language-like data via the discovery of full and partial matches between patterns, via the creation of referential linkages between patterns, and via heuristic search through the abstract space of alternative grammars (Section \ref{us_learning_in_sp_system_section}), it envisaged that, with facilities for the representation and processing of 2D patterns (Section \ref{patterns_and_symbols_section}), the SP system may build visual grammars from a robot's visual input \cite[Section 5]{sp_vision}.

As with other kinds of knowledge, grammars of that kind may include class hierarchies, part-whole hierarchies, and other kinds of knowledge structure, with seamless integration of different structures, as outlined in Section \ref{ma_knowledge_section} (see also \cite[Section 5.5]{sp_vision}).

It appears that, in deriving structures from visual input, there are important roles for binocular vision \cite[Section 5.1]{sp_vision} and for objects in motion \cite[Sections 5.2 and 5.3]{sp_vision}.

\subsection{How a Robot May Build 3D Models of Objects, of Itself, and of Its Environment}\label{building_model_of_world_section}

When it has been generalised for the representation and processing of 2D patterns, the multiple alignment framework may be applied in creating models of objects (including robots), and of a robot's environment. This is described fairly fully in \cite[Sections 6.1 and 6.2]{sp_vision} and summarised here.

The basic idea is that partially-overlapping images (from the robot's eyes) may be stitched together to create a coherent whole, in much the same way that partially-overlapping digital photographs may be stitched together to create a panorama. This is a relatively simple application of the multiple alignment concept, where a section at the end of one pattern matches a section at the beginning of another pattern. The system's ability to find good partial matches means that it should not be unduly disturbed by errors or distortions, provided they are not too great.

\begin{figure}[!htbp]
\centering
\includegraphics[width=0.4\textwidth]{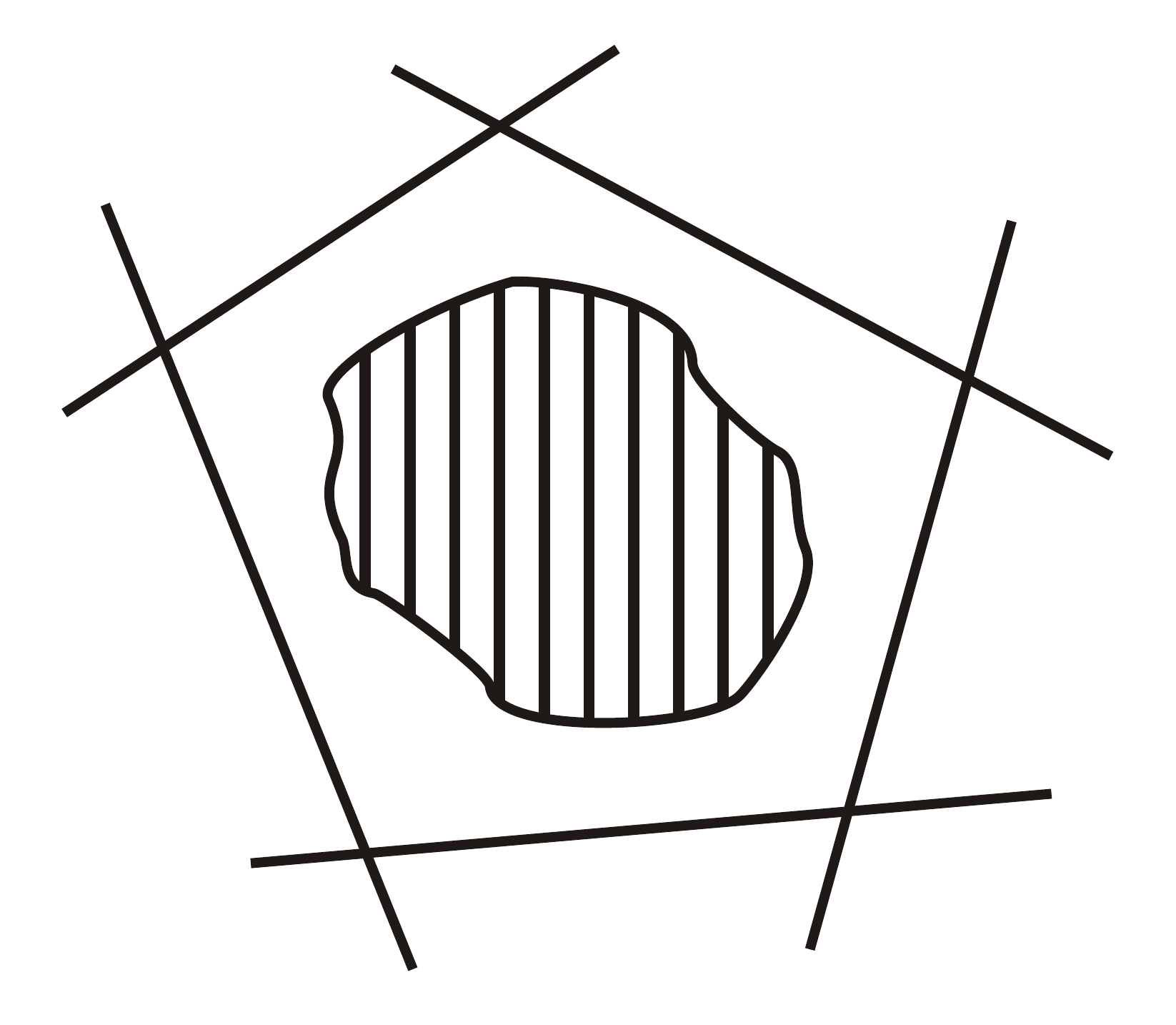}
\caption{Plan view of a 3D object, with each of the five lines around it representing a view of the object, as seen from the side. Reproduced from Figure 11 in \cite{sp_vision}, with permission.}
\label{3d_object_views_figure}
\end{figure}

Figure \ref{3d_object_views_figure} shows schematically how this idea may be applied to create a full or partial model of a 3D object. It is envisaged that overlapping views around the object may be stitched together to create the model.

We may have confidence in the feasibility of creating spatial models via this kind of mechanism because there are already commercially-available systems for the creation of digital 3D models from photographs. Examples include ``Big Object Base'' (\href{http://bit.ly/1gwuIfa}{bit.ly/1gwuIfa}), ``Camera 3D'' (\href{http://bit.ly/1iSEqZu}{bit.ly/1iSEqZu}), and ``PhotoModeler'' (\href{http://bit.ly/MDj70X}{bit.ly/MDj70X}).

Of course, an autonomous robot is itself a 3D object so we may suppose that similar principles may be applied to create a digital model of the robot itself. Bearing in mind that a robot would not normally be able to see all parts of itself, we may suppose that any model of itself that it builds from visual information is likely to partial, perhaps an adjunct to information about its organisation and workings that it gains from internal sensors.

Taking a broader view, the SP system may build a 3D model of a robot's environment in much the same way that Google's ``Streetview'' is built from overlapping pictures.\footnote{See also Google's ``Project Tango'', {\em Wikipedia}, \href{http://bit.ly/1mR8cM6}{bit.ly/1mR8cM6}, retrieved 2014-08-27.} The main difference between how Streetview models are constructed and how the SP system works, is that, with the SP system, images would be compressed via class hierarchies, part-whole hierarchies, and so on, as outlined in Section \ref{ma_knowledge_section}.

Building a 3D model of a robot's environment via multiple alignment is a potential solution to the ``mapping'' part of ``simultaneous localization and mapping'' (SLAM).\footnote{See, for example, ``Simultaneous localization and mapping'', {\em Wikipedia}, \href{http://bit.ly/1ikQTRR}{bit.ly/1ikQTRR}, retrieved 2014-08-06} The SP system may also provide a solution to the ``localization'' problem in SLAM via its capabilities in recognition (Section \ref{pattern_recognition_section}): using the multiple alignment system, it may compare its current visual input with its stored map of its surroundings and thus determine where it is, much as people appear to do.\footnote{A potentially interesting area of investigation is how these ideas may relate to the concept of a ``place cell'' ({\em Wikipedia}, retrieved 2014-10-07, \href{http://bit.ly/1vM6p26}{bit.ly/1vM6p26}) and associated neurophysiological evidence.}

\subsection{Interactions and Other Regularities}\label{interactions_regularities_section}

To operate effectively in the world, an autonomous robot needs some understanding of: how its parts work together; the impact of its actions on its surroundings; the impact of its surroundings on itself; and many other regularities in the world. Examples include: limitations on how the robot's limbs may be arranged; how pushing a mug or tumbler over the edge of a table means that it will normally fall and, very often, break when it hits the floor; how striking a bell will produce one kind of sound while striking a lump of wood will produce another; the damage and pain that can be caused by walking into a brick wall or going too close to a fire; sensations of strain if the robot tries to lift something that is too heavy; the way night follows day and day follows night; and the way that puddles evaporate, especially when the sun shines.

In broad terms, these many regularities in the world and in the way that a robot may interact with the world may be learned in the same way that words and other syntactic patterns may be learned from samples of language: candidate patterns may be identified via multiple alignment (Section \ref{creating_candidate_patterns_section}) and `good' grammars (collections of SP patterns) may be compiled via heuristic search (Section \ref{compiling_alternative_grammars_section}).

The key difference between learning syntactic structures from a one-dimensional sample of text and learning the kinds of regularities considered here is that the latter comprise arbitrary combinations of sights, sounds, sensations from tactile and proprioceptive receptors, tastes, smells, and so on; and that, very often, these inputs may be seen to comprise parallel streams of information in which two or more events may occur simultaneously.

This difference between learning from a one-dimensional stream of information and learning from parallel streams of information may be accommodated with three refinements of the SP70 model:

\begin{itemize}

\item {\em Represent parallel streams of information with 2D patterns}. If, as envisaged, the SP system is generalised to work with 2D patterns, such patterns may serve to represent parallel streams of information, as described in Appendix \ref{parallel_streams_of_information_section}.

\item {\em Generalise the sequence alignment process to the matching of 2D patterns}. The method for finding good full and partial matches between patterns (Section \ref{mup_section}) may be generalised to create an equivalent capability with 2D patterns.

\item {\em Generalise the process for building multiple alignments to accommodate 2D patterns}. The process for building multiple alignments should provide for the inclusion of 2D patterns.

\end{itemize}

With regard to the second point, a key feature of the existing process for sequence alignment is that it can find good matches between sequences despite interpolations of non-matching symbols in either or both sequences, as illustrated in Figure \ref{flexible_matching_figure}. It is envisaged that this kind of capability may be generalised to two dimensions and will then provide, in conjunction with heuristic search for good grammars (Section \ref{compiling_alternative_grammars_section}), a powerful means of finding recurrent patterns in parallel streams of information (or images) despite noise in the data.

\begin{figure*}[!htbp]
\fontsize{08.00pt}{09.60pt}
\centering
{\bf
\begin{BVerbatim}
0           a b I N     c d F O R e f M   g h A T       I     i j O N        k l m 0
                | |         | | |     |       | |       |         | |
1 X 1 n o p     I N q r     F O R     M s     A T t u v I w x     O N y z #X       1
\end{BVerbatim}
}
\caption{A multiple alignment produced by the SP computer model showing how two instances of the pattern `\texttt{I N F O R M A T I O N}' may be detected despite the interpolation of non-matching symbols throughout both instances.}
\label{flexible_matching_figure}
\end{figure*}

\subsection{Exploration, Play, and the Learning of Minor Skills}\label{exploration_play_section}

A robot that spends most of its existence sitting quietly in a cupboard might never know that night follows day or that puddles evaporate. In order to learn these humdrum features of the world---things that may seem too elementary and obvious to be taught (Section \ref{forms_of_learning_section})---our knowledge-hungry robot must explore. Robots that aspire to human-like intelligence must get about in the world and have varied experiences, as people do.

In a similar way, a robot that does not move or interact with the world cannot learn how its own body works, or the effects of its actions on its surroundings, or the impact of its surroundings on itself. Like people, especially children, robots must play. This is not merely a time-wasting indulgence, it is an essential part of learning the recurrent patterns that govern how people or robots may function in the world, including minor skills.

Although introspection can be a poor guide to mental processes, something may be gained from considering our experience in learning simple skills such as the childhood favourite: wiggling one's ears. In learning something like that, it is initially hard to know what to do, but then things gradually take shape. What seems to be happening is that we are trying out different combinations and sequences of muscle contractions and of relaxations of muscles to discover what works. Those combinations and sequences, with feedback about what does and does not work, may be represented as parallel streams of information in 2D patterns, as described in Appendix \ref{parallel_streams_of_information_section}.

\subsection{Learning a Major Skill Via Practice}\label{learning_skills_section}

In addition to the knowledge and skills that may be gained via exploration and play, an autonomous robot, like a person, is likely to need more elaborate skills.

As a rough generalisation in that connection, the current generation of robots often have the benefit of artificial aids or simplifications of tasks, their skills are normally programmed by people, case by case, and they are generally deficient in human-like abilities to learn new skills for themselves. For example, the ``Deep Green'' pool-playing robot \cite{greenspan_etal_2008} has the benefit of an overhead camera giving a birds-eye view of the pool table, it is programmed with relevant knowledge of geometry and of Newtonian physics, and it does not aspire to learn such things as how to make bread or how to play tennis.

Although there is more to games like pool, billiards and snooker than merely potting balls, the process of potting a ball is a significant challenge for human players, especially with the large table of billiards and snooker. Here we consider how that skill, and related skills, may be learned via the SP system under the same conditions as human players, extending the discussion in \cite[Section 6.3]{sp_benefits_apps}.

When a human player pots a ball, he or she strikes the cue ball with the cue, aiming to send the cue ball towards the target ball to strike it in such a way that the target ball is propelled towards a pocket and falls into it. Skilled players may influence the behaviour of the cue ball or the target ball or both by imparting spin to the cue ball (by striking it away from the centre). And in the course of a game, skilled players may combine the potting of the target ball with measures to ensure that other balls finish their movements in positions that are advantageous for the next shot. Indeed, in snooker for example, the intention with some shots is not to pot any ball but merely to move balls into positions that make things difficult for one's opponent (ie to create a `snooker'). Since that kind of shot requires the same kinds of skills as are needed for potting a ball, our discussion will be about both kinds of shot. To simplify things in the discussion below, we shall ignore the fact that, in potting a ball, most human players will take advantage of depth perception via binocular vision.

Unlike the Deep Green robot, the human player depends mainly on a view along the cue towards the cue ball and beyond, without a birds-eye view of the table via a ceiling-mounted camera.\footnote{Deep Green actually has both those views, and human players may walk round the table to get different views. The key point here is that the human player does not normally get the birds-eye view.} Unlike the Deep Green robot, most human players have little or no formal knowledge of Newtonian physics or geometry, and even if they had, the absence of a birds-eye view would make such knowledge less useful than it is for the robot. It seems likely that the performance of skilled human players in potting balls, in the use of spin, and in the positioning of balls, has little to do with physics or geometry and is much more to do with extensive experience via thousands of hours of practice.

The SP system may support that kind of learning in an autonomous robot as outlined in what follows. For each shot:

\begin{itemize}

\item {\em Before the shot: parse visual input}. Before the given shot is taken, the robot may assimilate information about the configuration of the table. In one or more views, the robot will see the cue ball, the target ball, other balls that may be on the table, and the target pocket (when the intention is to pot a ball). Each of those views may be parsed into its parts and sub-parts, much as in the parsing of natural language (Section \ref{parsing_nl_section}, \cite[Section 4]{sp_vision}). Each parsing is a multiple alignment from which an encoding may be derived as indicated in Section \ref{ma_section}. There is potential for a 3D model to be derived from multiple views, as outlined in Section \ref{building_model_of_world_section}.

\item {\em During the shot: record actions and effects}. As the shot is taken, a record may be kept of such things as contractions by the muscles of the robot, the speed and direction of the cue as it strikes the cue ball, the point on the cue ball that is struck by the cue (relative to the centre of the cue ball), tactile feedback from the impact of the cue on the cue ball, auditory feedback from impacts of various kinds, and observations of movements of the balls, perhaps analysed as outlined in \cite[Section 5.3]{sp_vision}. If the target ball has been successfully potted, that event will be recorded too. As in the learning of interactions and other regularities (Section \ref{interactions_regularities_section}), there will be parallel streams of information which may be recorded using 2D patterns, as outlined in Appendix \ref{parallel_streams_of_information_section}.

\item {\em At the end of the shot: parse visual input}. At the end of the shot, as at the beginning, the robot may assimilate information about the configuration of the table via one or more views, with multiple alignment as the means of parsing each view. As before, there is potential for a 3D model to be derived from multiple views.

\end{itemize}

For each shot, the overall result will be a set of SP patterns that record: 1) The configuration of the playing table before the shot; 2) Actions and effects as the shot is taken; and 3) The configuration of the table at the end of the shot---each one recorded in compressed form. Each such set of patterns may be regarded as one row in a database of configurations, actions and effects in the potting of balls. As the robot practices, it will build up the database, which may eventually become very large.

Although the three elements of each row will, individually, be compressed, there is likely to be scope for further abstraction and compression of the whole database via the discovery of redundancies across the rows of the database.

The whole database may serve as a guide for future shots. For any such shot, the robot may find the row that best matches the initial configuration and the desired outcome, and it may then perform the corresponding actions.

\subsection{Learning Via Demonstration}\label{learning_via_demonstration_section}

A popular approach to the training of robots is simply to demonstrate directly what the robot is to do, normally by guiding one or more of the robot's limbs through the required motions.\footnote{See, for example, ``This robot could transform manufacturing'', {\em MIT Technology Review}, 2012-09-18, \href{http://bit.ly/1nbnJfv}{bit.ly/1nbnJfv}; ``Robots that learn through repetition, not programming'', {\em MIT Technology Review}, 2014-09-22, \href{http://bit.ly/1shxuLk}{bit.ly/1shxuLk}.}. This is much simpler than learning by imitation (Section \ref{forms_of_learning_section}) because it by-passes the relatively complex perceptual skills needed in the latter case.

With this kind of learning, the SP system may have a useful role to play by compressing the information gathered about the demonstrated motion of the robot's limb or limbs. It has the potential, for example, to identify repeating elements in the demonstrated motion and to abstract them as distinct subroutines. More generally, it may create a grammar for the required motion including hierarchical structures, classes of structure, and so on.

A potential benefit of this kind of information compression is generalisation of the demonstrated motion, without over-generalisation, as described in the next subsection. This may help to provide a degree of flexibility in the robot's actions. If sensory inputs are provided in conjunction with the demonstrated motion, the robot may also develop some ability to adapt to changes in the required task.

\subsection{Generalisation, Correction of Over-Generalisations, and Dirty Data, in the Learning of Non-Verbal Behaviour}\label{generalisation_section}

As we have seen (Section \ref{computer_models_of_language_learning_section}), the principle of minimum length encoding provides a neat explanation of three aspects of language learning: 1) How we can generalise our knowledge from the finite sample of utterances which is the basis for our learning to the infinite range of utterances in the target language, {\bf L}; 2) How over-generalisations may be corrected without error correction by a teacher, or anything equivalent; and 3) How people can develop a strong sense of what utterances do and do not belong in {\bf L} when, in the vast majority of cases, there are errors in the sample of language that is the basis for our learning.

Naturally, these principles would apply to language learning by robots (Section \ref{learning_linguistic_knowledge_section}). But it appears that the principles may also apply to the learning of non-verbal behaviour by a person or an autonomous robot:

\begin{itemize}

% Choose a better example

\item {\em Grammars for non-verbal behaviour}. A set of skills like those required for cooking a meal may be regarded as a kind of non-verbal language. In view of the generality of the multiple alignment concept and the learning mechanisms in the SP system, it seems likely that grammars for that kind of non-verbal behaviour, and others, may be learned in much the same way as grammars for natural language.

\item {\em Generalisation without over-generalisation}. While the grammar for cooking a meal would be derived from specific experiences, it may generate an infinite range of possible meals. But it is not totally general and will not, for example, enable a person or robot to play poker.

\item {\em Learning from dirty data}. In learning the skills needed for cooking a meal, we can be successful despite the likelihood that people that we learn from may make mistakes and cook books are rarely free of errors.\footnote{In my own experience, I once learned to program a computer in assembly language using an instruction manual that was riddled with errors. The fact that much of the manual was correct made it possible to identify the errors and work around them.}

\end{itemize}

\subsection{Cutting the Cost of Learning}\label{cutting_learning_costs_section}

A familiar feature of human learning is that it can be costly in terms of time and money. Even with a natural talent, it can take a great deal of practice to become skilled in playing a musical instrument or in sports. People spend much time in education and training, mainly when they are young but also later in life, and schools, colleges and teachers all cost money.

An interesting possibility with autonomous robots and other artificial systems is that much of this cost may be avoided. This is because knowledge or skills that have been built up by one robot may be downloaded easily and transferred to any number of other robots. Naturally, this works best when the several robots are identical but otherwise there is potential for adjustments to be made via learning when the recipient robots are similar to the donor robot but not exactly the same.

\section{Conclusion}

The {\em SP theory of intelligence} and its realisation in the {\em SP machine} may facilitate the development of autonomous robots: by increasing the computational efficiency of computers; by facilitating the development of human-like versatility in intelligence; and likewise for human-like adaptability in intelligence.

With regard to the first problem, the SP system has potential for substantial gains in computational efficiency, with corresponding cuts in energy consumption and in the bulkiness of computing machinery: by reducing the size of data to be processed; by exploiting statistical information that the system gathers as an integral part of how it works; and via an updated version of Hebb's concept of a ``cell assembly''.

In the quest for human-like versatility in intelligence, the SP system has strengths in several areas including unsupervised learning, natural language processing, pattern recognition, information retrieval, several kinds of reasoning, planning, problem solving, and more, with seamless integration amongst structures and functions.

The SP system may also promote human-like adaptability in intelligence via its strengths in unsupervised learning. As has been discussed, these capabilities may underpin one-trial learning, the learning of natural language, learning to see, building 3D models of objects and of a robot's surroundings, learning how a robot interacts with its environment and other regularities, learning minor skills via exploration and play, learning major skills, and learning via demonstration. Associated issues that have been discussed include: learning from parallel streams of data; generalisation, correction of over-generalisations, and learning from dirty data; how to cut the cost of learning; and reinforcements and motivations.

\sloppy Although it is likely that autonomous robots will require a non-von revolution---perhaps along the lines of SP-neural---there is plenty that can be done via modelling with von-Neumann-style supercomputers to explore the potential of new architectures. One such possibility is research with a high-parallel version of the SP machine, as outlined in Section \ref{computer_model_and_sp_machine_section}. This would be a means for researchers everywhere to explore what can be done with the system and to create new versions of it.

\appendices

\section{How Symbols May Be Identified}\label{identification_of_symbols_section}

This appendix and the ones that follow consider issues that are relevant to discussions elsewhere in this paper: how symbols may be identified; how quantification can or should be accommodated in the SP system; and how the SP system may represent and process parallel streams of information.

In the SP system, in the processing of things like text, it is a straightforward matter to equate individual characters, or whole words, with SP symbols. But identifying SP symbols is less easy with things like images, or recordings of speech or music:
\begin{itemize}

\item {\em Images}. One possibility with images is to treat each pixel as an SP symbol, where each pixel may be matched in an all-or-nothing manner with any other pixel. Another possibility is, via some conventional pre-processing, to identify low-level features in images such as lines and angles and to treat such features as SP symbols. How that kind of thing may be done within the SP framework is discussed in \cite[Section 3]{sp_vision}.

\item {\em Speech}. As with images, it may be possible to treat the lowest-level elements as SP symbols. Alternatively, features such as white noise (from fricative consonants), formants, formant ratios, and formant transitions, may be identified via pre-processing and treated as being SP symbols.

\item {\em Music}. A relatively straightforward analysis would equate individual notes (including notes within chords) as being SP symbols. As with images and speech, it may be necessary on relatively short timescales to use conventional pre-processing (Fourier analysis and the like) to isolate individual notes within a stream of music.

\end{itemize}

Of course, hybrid solutions, using conventional pre-processing in conjunction with the SP system, are not as theoretically neat as when everything is done via the SP system, but they may be justified as short-term expedients. On longer timescales, it would probably make better sense to try to avoid the relative complexity of such solutions. In the processing of images, for example, there is potential for the SP system to isolate features such as lines and angles \cite[Sections 3.3 and 3.4]{sp_vision}.

\section{Quantification in the SP System}\label{quantification_section}

In understanding any kind of skilled activity---playing tennis, cooking a meal, and so on---or creating robots that have such skills, it seems natural to measure things like the strength with which a tennis player hits a ball, and to express the measurements with numbers. But the SP system, in itself, makes no provision for quantification, neither analogue nor digital.\footnote{It is true that each SP pattern has an associated frequency of occurrence (Section \ref{outline_of_sp_system}) but that is for internal use and not for representing something like the strength with which a tennis player hits a ball.} At the lowest level in its knowledge, it deals with symbols which have no intrinsic meaning, numerical or otherwise (Section \ref{outline_of_sp_system}). As noted in Section \ref{patterns_and_symbols_section}, any meaning associated with a given SP symbol or combination of symbols must be expressed using other SP symbols, or external equivalents; and there is just one valid operation with an SP symbol: to match it with another SP symbol to determine whether they are the same or different.

In broad terms, there are three main ways in which quantification may be accommodated in the SP system:

\begin{itemize}

\item {\em Via densities of symbols}. The densities of different categories of symbols may serve to express quantities, in much the same way that the densities of black and white pixels may represent different shades of grey in a black and white photograph, at least as they used to be \cite[Section 2.2.3]{wolff_2006}.

\item {\em Via the rules of arithmetic}. In principle, the SP system may express values as numbers and perform arithmetic operations if it is supplied with SP patterns representing Peano's axioms and other knowledge about arithmetic. This has not yet been explored in any depth. There is relevant discussion in \cite[Chapter 10]{wolff_2006}.

\item {\em Don't do it}. The simplest option is to avoid quantification altogether. This may not be as silly as it sounds, as suggested in the following discussion.

\end{itemize}

Although it may seem natural to quantify the operations of a robot and to represent quantities with numbers, those assumptions may carry with them an unspoken and possibly unjustified belief that increasing the size of a given value in the robot would normally increase the impact of that value in the robot's environment. For example, it seems obvious, and is probably true, that if an industrial robot fails to bend a metal bar with medium pressure, stronger pressure is likely to succeed.

But the assumption of a linear or monotonic relationship between variables is often wrong. If we hold an egg in our hand with a grip that is too weak, we may drop it. But if our grip is too strong, the egg may be crushed. Holding an egg works best with a grip that is neither too weak nor too strong. Returning to our tennis player example, strong blows to the ball may score points on many occasions but with a drop shot, for example, a light touch is required.

In general, any person or robot must keep an open mind in learning how things work, without presuppositions about linear or quasi-linear relationships between variables. In that case, it is an advantage rather than a handicap if values like {\em weak}, {\em medium}, and {\em strong} are represented with SP symbols without any presumption of a quantitative relationship amongst them, much as one would assume for values like {\em red}, {\em sweet}, and {\em curly}. Elsewhere in this paper we shall assume that all values that provide input for a robot's learning are represented with standard SP symbols, without any quantitative meaning.

\section{How the SP System May Represent and Process Parallel Streams of Information}\label{parallel_streams_of_information_section}

In the development to date of the SP theory and the SP computer model, the main focus has been on one-dimensional patterns and what can be done with them. This can work well with some kinds of information, as can be seen in Figures \ref{class_part_plant_figure} and \ref{parsing_figure}. In each case, the patterns in the multiple alignment may be merged (unified) to create a single 1D pattern.

In examples like Figure \ref{parsing_figure}, the left-to-right ordering of symbols represents the time ordering of words and other structures in natural language. This is OK for a one-dimensional stream of information like ordinary text but is not satisfactory when there are two or more streams of information in parallel. Here are some examples:

\begin{itemize}

\item {\em Speech}. In both the production of speech and in the acoustic signal of speech, there are normally several things going on in parallel. When we speak, there is simultaneous activity in our lips, tongue, cheeks, and breathing; while in terms of acoustics, elements of speech such as vowel sounds may be distinguished, one from another, by configurations of simultaneously-occurring spectral peaks or formants.

\item {\em Spelling rules}. Notwithstanding the `whole word' doctrine in the teaching of reading, it is widely recognised that, with English at least, skilled readers know many associations between configurations of letters and corresponding sounds: `\texttt{th}' = \texttheta~or \dh; `\texttt{ch}' = \textteshlig; `\texttt{ay}' = e\textsci; and so on.\footnote{In each case, the sounds, to the right, are represented with symbols from the International Phonetic Alphabet.} In terms of before-and-after relationships, it makes most sense to say that a spelling pattern and its sound value occupy the same time slot.

\item {\em Music}. In music, especially orchestral music and music for the piano or organ, it is normal for two or more notes to be played at the same time.

\item {\em Natural language and its meanings}. The surface forms of spoken or written language, and the meanings of those forms, may be seen as parallel streams of information (see also Sections \ref{nl_parallel_streams_section} and \ref{learning_linguistic_knowledge_section}).

\item {\em Robots and their surroundings}. In considering the information that an autonomous robot needs to process, there are normally several streams of information running in parallel, in two main areas:

\begin{itemize}

\item {\em The robot's environment}. As with people, there would normally be several things happening at the same time. In a typical office, for example, there would be people talking, phones ringing, people coming and going, people working on keyboards, taking refreshment, doing photocopying, and so on.

\item {\em The robot's workings and its impact on its surroundings}. In any robot of reasonable complexity, there would be signals going to the robot's `muscles', and, via sensors of various kinds, there would be information about the robot's surroundings, information about the internal workings of the robot, and feedback about the effects of the robot's actions on its surroundings.

\end{itemize}

\end{itemize}

\subsection{Representing Parallel Streams of Information With Two-Dimensional Patterns}\label{2d_patterns_section}

\begin{figure}[!htbp]
\centering
\includegraphics[width=0.45\textwidth]{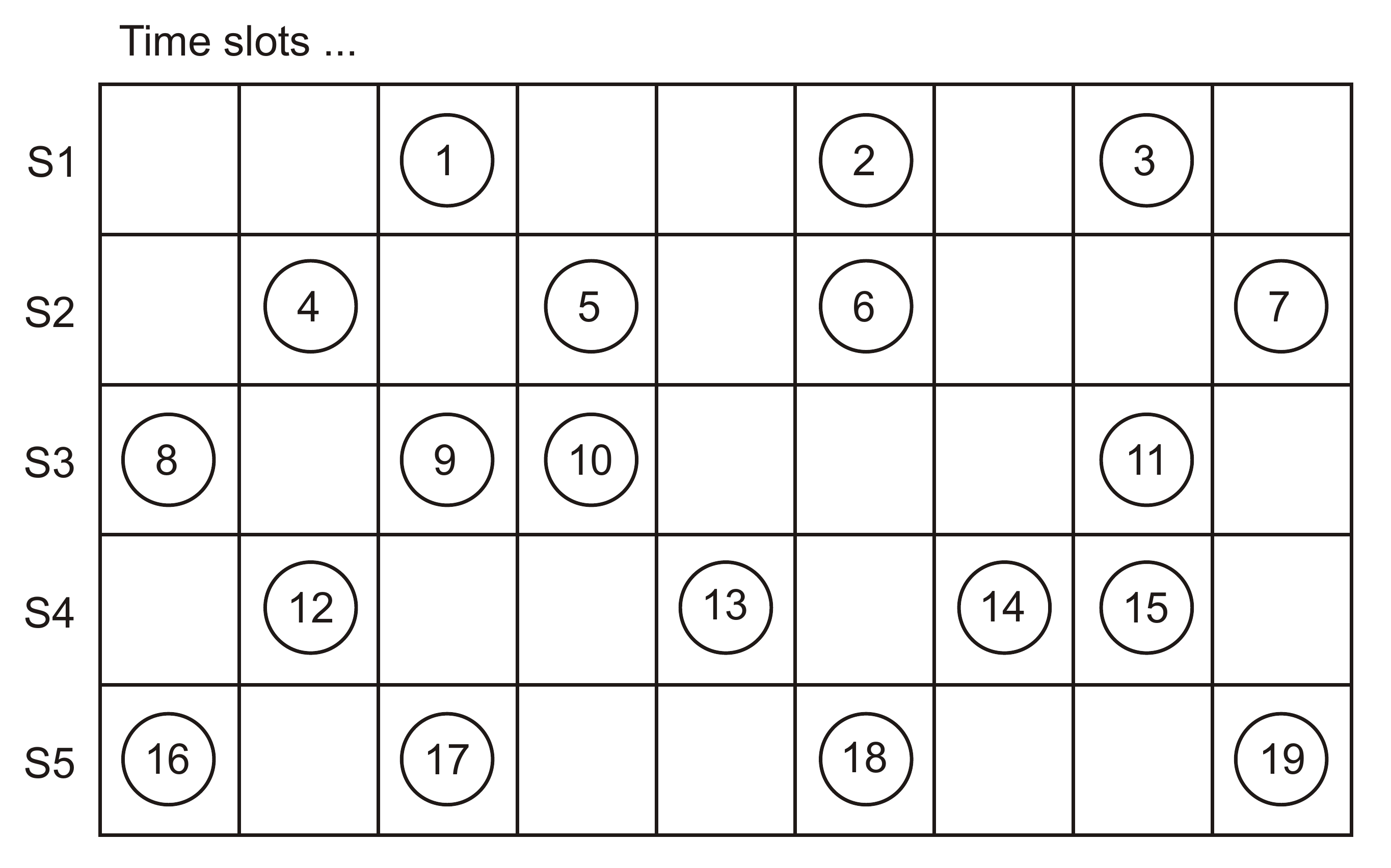}
\caption{An example of a 2D SP pattern showing how it may be used to represent parallel streams of information. Each row (S1 to S5) represents a stream of information and each column is a time slot. Each numbered circle is an SP symbol.}
\label{parallel_streams_figure}
\end{figure}

How should the SP system accommodate parallel streams of information? The most straightforward answer seems to be to take advantage of what is in any case envisaged for development within the system: SP patterns in two dimensions. Such patterns were originally conceived, and are still seen, as a vehicle for the representation and processing of images \cite{sp_vision}. But they may also serve in the representation and processing of parallel streams of information, as illustrated in Figure \ref{parallel_streams_figure}.

In this example, any given SP symbol (shown as a numbered circle) may be assigned to a time slot (one of the columns in the figure) in one of five streams of information (shown as rows S1 to S5 in the figure).

Aspects of this proposal are described and discussed in the subsections that follow.

\subsection{Streams of Information}\label{streams_section}

The reason for introducing the notion of `streams' of information is to try to achieve some consistency across different configurations of symbols, and the reason for that is to facilitate the discovery of patterns that match each other, as indicated in Section \ref{mup_section} and amplified in Section \ref{us_learning_in_sp_system_section}. For example, it is easier to recognise that one musical chord is the same as another if symbols for the constituent notes are arranged in order of their pitch (as they normally are in musical notations) than if there is a haphazard ordering of symbols from one instance of a chord to another. Like the different levels in a musical stave, streams of information provide a means of keeping things in order.

It seems likely that some streams of information will be for relatively concrete attributes like pitch, while others will be for relatively abstract attributes arising from the fact that some symbols may serve as references to other structures, as outlined in Sections \ref{ma_knowledge_section} and \ref{references_and_abstractions_section}.

\subsection{Extended Events}\label{extended_events_section}

Figure \ref{parallel_streams_figure} may give the impression that, in the proposed scheme for representing parallel streams of information, every symbol represents some kind of short event occupying a single time slot. But in music for example, individual notes vary in their duration and some, such as the drone of a bagpipe, may be held for extended periods.

In the proposed scheme, any event that is longer than a single time slot may be represented with a pair of symbols, one marking the start of the event and the other marking when it ends. For example, the symbol `10' in stream S3 in Figure \ref{parallel_streams_figure} may be a mark for the beginning of a note, while the symbol `11' in the same stream may mark when the note ends.

\subsection{References and Abstractions}\label{references_and_abstractions_section}

As mentioned in Section \ref{streams_section}, 2D patterns representing parallel streams of information may contain symbols that serve as references to other structures. For example, where a portion of music is repeated in different parts of a musical composition, the first instance may be marked with a `start' symbol and an `end' symbol, and then copies of those two symbols may serve to represent later instances without the need to repeat all the detail.

As outlined in Section \ref{ma_knowledge_section}, these kinds of linkages between patterns may serve to define classes and subclasses of structure, parts and sub-parts, and several other kinds of knowledge, with seamless integration of different structures.

\subsection{Processing of Parallel Streams of Information}

It is envisaged that 2D patterns representing parallel streams of information will be processed in the same way as other SP patterns: via processes for finding good full and partial matches between patterns and via processes for the building of multiple alignments. As with the processing of 2D patterns representing other kinds of data such as images, it will be necessary to generalise the processes for finding good full and partial matches between patterns and for building multiple alignments so that they work with 2D patterns. There is further discussion in Section \ref{interactions_regularities_section}.

\bibliographystyle{plain}
% \bibliography{latex_references}

\begin{thebibliography}{10}

\bibitem{brown_1989}
C.~Brown.
\newblock {\em My Left Foot}.
\newblock Mandarin, London, 1989.

\bibitem{chomsky_1959}
N.~Chomsky.
\newblock A review of {B}.~{F}.~{S}kinner's ``{V}erbal {B}ehavior''.
\newblock {\em Language}, 35(1):26--58, 1959.
\newblock Re-published, with a preface by N. Chomsky, in L. A. Jakobovits and
  M. S. Miron (eds.), {\em Readings in the Psychology of Language},
  Prentice-Hall, 1967, pp.~142--143.

\bibitem{gold_1967}
M.~Gold.
\newblock Language identification in the limit.
\newblock {\em Information and Control}, 10:447--474, 1967.

\bibitem{greenspan_etal_2008}
M.~Greenspan, J.~Lam, M.~Godard, I.~Zaidi, S.~Jordan, W.~Leckie, K.~Anderson,
  and D.~Dupuis.
\newblock Toward a competitive pool-playing robot.
\newblock {\em Computer}, 41(1):46--53, 2008.

\bibitem{hebb_1949}
D.~O. Hebb.
\newblock {\em The Organization of Behaviour}.
\newblock John Wiley \& Sons, New York, 1949.

\bibitem{herculano-houzel_2012}
S.~Herculano-Houzel.
\newblock The remarkable, yet not extraordinary, human brain as a scaled-up
  primate brain and its associated cost.
\newblock {\em Proceedings of the National Academy of Sciences of the United
  States of America}, 109(Supplement 1):10661--10668, 2012.

\bibitem{kelly_hamm_2013}
J.~E. Kelly and S.~Hamm.
\newblock {\em Smart machines: {IBM}'s {W}atson and the era of cognitive
  computing}.
\newblock Columbia University Press, New York, 2013.

\bibitem{lenneberg_1962}
E.~H. Lenneberg.
\newblock Understanding language without the ability to speak.
\newblock {\em Journal of Abnor¬mal and Social Psychology}, 65:419--425, 1962.

\bibitem{li_vitanyi_2009}
M.~Li and P.~Vit\'{a}nyi.
\newblock {\em An Introduction to Kolmogorov Complexity and Its Applications}.
\newblock Springer, New York, 2009.

\bibitem{quine_1960}
W.~V.~O. Quine.
\newblock {\em Word and Object}.
\newblock MIT Press, Cambridge, MA., 1960.

\bibitem{rissanen_1978}
J.~Rissanen.
\newblock Modelling by the shortest data description.
\newblock {\em Automatica-J, {IFAC}}, 14:465--471, 1978.

\bibitem{solomonoff_1964}
R.~J. Solomonoff.
\newblock A formal theory of inductive inference. {P}arts {I} and {II}.
\newblock {\em Information and Control}, 7:1--22 and 224--254, 1964.

\bibitem{turing_1950}
A.~M. Turing.
\newblock Computing machinery and intelligence.
\newblock {\em Mind}, 59:433--460, 1950.

\bibitem{wallace_boulton_1968}
C.~S. Wallace and D.~M. Boulton.
\newblock An information measure for classification.
\newblock {\em Computer Journal}, 11(2):185--195, 1968.

\bibitem{wolff_1988}
J.~G. Wolff.
\newblock Learning syntax and meanings through optimization and distributional
  analysis.
\newblock In Y.~Levy, I.~M. Schlesinger, and M.~D.~S. Braine, editors, {\em
  Categories and Processes in Language Acquisition}, pages 179--215. Lawrence
  Erlbaum, Hillsdale, NJ, 1988.
\newblock See \href{http://bit.ly/ZIGjyc}{bit.ly/ZIGjyc}.

\bibitem{wolff_1993}
J.~G. Wolff.
\newblock Computing, cognition and information compression.
\newblock {\em AI Communications}, 6(2):107--127, 1993.
\newblock See \href{http://bit.ly/XL359b}{bit.ly/XL359b}.

\bibitem{wolff_medical_diagnosis}
J.~G. Wolff.
\newblock Medical diagnosis as pattern recognition in a framework of
  information compression by multiple alignment, unification and search.
\newblock {\em Decision Support Systems}, 42:608--625, 2006.
\newblock See \href{http://bit.ly/XE7pRG}{bit.ly/XE7pRG}.

\bibitem{wolff_2006}
J.~G. Wolff.
\newblock {\em Unifying Computing and Cognition: the {SP} Theory and Its
  Applications}.
\newblock CognitionResearch.org, Menai Bridge, 2006.
\newblock {ISBN}s: 0-9550726-0-3 (ebook edition), 0-9550726-1-1 (print
  edition). Distributors, including Amazon.com, are detailed on
  \href{http://bit.ly/WmB1rs}{bit.ly/WmB1rs}.

\bibitem{wolff_sp_intelligent_database}
J.~G. Wolff.
\newblock Towards an intelligent database system founded on the {SP} theory of
  computing and cognition.
\newblock {\em Data \& Knowledge Engineering}, 60:596--624, 2007.
\newblock See \href{http://bit.ly/Yg2onp}{bit.ly/Yg2onp}.

\bibitem{sp_extended_overview}
J.~G. Wolff.
\newblock The {SP} theory of intelligence: an overview.
\newblock {\em Information}, 4(3):283--341, 2013.
\newblock See \href{http://bit.ly/1hz0lFE}{bit.ly/1hz0lFE}.

\bibitem{sp_vision}
J.~G. Wolff.
\newblock Application of the {SP} theory of intelligence to the understanding
  of natural vision and the development of computer vision.
\newblock 3(1):552, 2014.
\newblock See \href{http://bit.ly/1scmpV9}{bit.ly/1scmpV9}.

\bibitem{sp_big_data}
J.~G. Wolff.
\newblock Big data and the {SP} theory of intelligence.
\newblock {\em IEEE Access}, 2:301--315, 2014.
\newblock See \href{http://bit.ly/1jGWXDH}{bit.ly/1jGWXDH}.

\bibitem{sp_benefits_apps}
J.~G. Wolff.
\newblock The {SP} theory of intelligence: benefits and applications.
\newblock {\em Information}, 5(1):1--27, 2014.
\newblock See \href{http://bit.ly/1lcquWF}{bit.ly/1lcquWF}.

\end{thebibliography}

\end{document}